%% file: main.tex
\documentclass[letterpaper]{article} % DO NOT CHANGE THIS
\usepackage{aaai25}  % DO NOT CHANGE THIS
\usepackage{times}  % DO NOT CHANGE THIS
\usepackage{helvet}  % DO NOT CHANGE THIS
\usepackage{courier}  % DO NOT CHANGE THIS
\usepackage[hyphens]{url}  % DO NOT CHANGE THIS
\usepackage{graphicx} % DO NOT CHANGE THIS
\usepackage{natbib}  % DO NOT CHANGE THIS AND DO NOT ADD ANY OPTIONS TO IT
\usepackage{caption} % DO NOT CHANGE THIS AND DO NOT ADD ANY OPTIONS TO IT
\usepackage{booktabs}
\usepackage[ruled,vlined]{algorithm2e}
\usepackage{multirow}
\usepackage{subfigure}
\usepackage{amsthm,amsmath,amssymb}
\title{S$^2$DN: Learning to Denoise Unconvincing Knowledge for Inductive Knowledge Graph Completion}
\author{
    Tengfei Ma\textsuperscript{\rm 1},
    Yujie Chen\textsuperscript{\rm 1},
    Liang Wang\textsuperscript{\rm 2,3},
    Xuan Lin\textsuperscript{\rm 4},
    Bosheng Song\textsuperscript{\rm 1}\thanks{Corresponding Author}, 
    Xiangxiang Zeng\textsuperscript{\rm 1}
}
\affiliations{
    \textsuperscript{\rm 1} College of Computer Science and Electronic Engineering, Hunan University, China \\
    \textsuperscript{\rm 2} NLPR, MAIS, Institute of Automation, Chinese Academy of Sciences \\
\textsuperscript{\rm 3} School of Artificial Intelligence, University of Chinese Academy of Sciences \\
    \textsuperscript{\rm 4} College of Computer Science, Xiangtan University, China \\   
}

\usepackage{bibentry}
\begin{document}

\maketitle

\begin{abstract}
Inductive Knowledge Graph Completion (KGC) aims to infer missing facts between newly emerged entities within knowledge graphs (KGs), posing a significant challenge. While recent studies have shown promising results in inferring such entities through knowledge subgraph reasoning, they suffer from (i) \textit{the semantic inconsistencies of similar relations}, and (ii) \textit{noisy interactions inherent in KGs} due to the presence of unconvincing knowledge for emerging entities. 
% This oversight hampers the prediction performance and robustness of the inductive KGC task. 
To address these challenges, we propose a \underline{S}emantic \underline{S}tructure-aware \underline{D}enoising \underline{N}etwork (S$^2$DN) for inductive KGC. Our goal is to learn adaptable general semantics and reliable structures to distill consistent semantic knowledge while preserving reliable interactions within KGs. Specifically, we introduce a semantic smoothing module over the enclosing subgraphs to retain the universal semantic knowledge of relations. We incorporate a structure refining module to filter out unreliable interactions and offer additional knowledge, retaining robust structure surrounding target links. Extensive experiments conducted on three benchmark KGs demonstrate that S$^2$DN surpasses the performance of state-of-the-art models. These results demonstrate the effectiveness of S$^2$DN in preserving semantic consistency and enhancing the robustness of filtering out unreliable interactions in contaminated KGs. Code is available at \url{https://github.com/xiaomingaaa/SDN}.
\end{abstract}

\input{intro}

\input{rel_work}
\input{preliminary}

\input{method}

\input{experiment}

\section{Conclusion}
% We introduced S$^2$DN to address the challenges posed by the semantic inconsistencies and inevitable noisy interactions in KGs for inductive KGC, emphasizing semantic consistency and structural reliability. Our experimental findings, conducted on benchmarks and custom-designed noisy KGs, show that S$^2$DN surpasses SOTA baselines by keeping relational semantic consistency and offering robust associations.
We introduced S$^2$DN to address the challenges posed by the semantic inconsistencies and inevitable noisy interactions in KGs for inductive KGC, emphasizing semantic consistency and structural reliability. Experimental results show that S$^2$DN surpasses SOTA baselines by keeping relational semantic consistency and offering robust associations. In future work, we will transfer S$^2$DN to more noise-sensitive domain applications such as biology.

% \section{Acknowledge}
\section{Acknowledgments}
The work was supported by the National Natural Science Foundation of China (62272151, 62122025, 62425204, 62450002, 62432011 and U22A2037), the National Science and Technology Major Project (2023ZD0120902), Hunan Provincial Natural Science Foundation of China (2022JJ20016), The science and technology innovation Program of Hunan Province (2022RC1099), and the Beijing Natural Science Foundation (L248013).

\bibliography{main}
\clearpage
\input{appendix}

\end{document}

%\maketitle

\tableofcontents

\section{Technical Appendix}

something

\appendix

\section{My Title}
my Appendix \thesection{}, starts the appendix\ldots

\section*{A starred section}
with some content

\section{Another section}
my Appendix \thesection{} is about\ldots 

\subsection{A subsection}

with some text

\nocite{*}
\bibliographystyle{plain}
\bibliography{bibs}

%% file: intro.tex
\section{Introduction}
% \begin{figure}[t]
% \centering
% \includegraphics[width=1\columnwidth]{figs/intro.pdf} % Reduce the figure size so that it is slightly narrower than the column. Don't use precise values for figure width. This setup will avoid overfull boxes.
% \caption{An example of noisy interactions and blurring relations. The query \textit{(Trump,live\_in,?)} asks which city \textit{Trump} lives in. The \textcolor{red}{red dashed lines} are noise that may bring an incorrect fact \textit{(Trump,live\_in, New York)}. In addition, the relations \textit{\textcolor{blue}{located\_in}} and \textit{\textcolor{blue}{lie\_in}} with semantic ``\textit{\textcolor{blue}{the location of}}'' may lead to inconsistent expressions by using different embeddings to represent them.} 
% \label{fig:intro}
% \end{figure}

Knowledge graphs (KGs) represent the relations between real-world entities as facts 
% in the form of \textit{(head entity, relation, tail entity)}, abbreviated as \textit{(h,r,t)}, 
providing a general way to store semantic knowledge~\citep{wang2017knowledge,zhang2023structure,KG_complete}. 
KGs have been successfully used in various applications, including recommendation systems~\citep{yang2023knowledge}, question answering~\citep{kg_qa,liu2023knowledge}, and drug discovery~\citep{lin2020kgnn,ma2022kg}. However, KGs often suffer from incompleteness~\citep{geng2023relational} and newly emerging entities~\citep{xu2022subgraph}. Thus, inductive knowledge graph completion (KGC), proposed to predict missing facts on unseen entities in KGs, has been a hot area of research~\citep{zhang2023adaprop,bai2023knowledge} and industry~\citep{du2021cogkr,KG_survey}. 
% To accomplish this goal, previous transductive KGC methods~\citep{bordes2013translating,sun2019rotate,KG_transductive_tnnls_1} learn specific embeddings for each entity and relation using complex transformations on vector spaces, which are adopted to predict the linking probability of unknown triples. But in the real-world KGs, many new entities are emerging constantly over time~\citep{xu2022subgraph} (e.g., new items in e-commerce). These transductive works have to re-train the whole KG under this scenario to introduce emerging entities, which is not infeasible in practice. To infer links between emerging entities, the inductive KGC task,  

To improve the generalization ability of the KGC task on unknown entities, some researchers propose rule-based methods~\citep{meilicke2018fine,yang2017differentiable,sadeghian2019drum}. These methods enable effective reasoning under emerging entities by mining common reasoning rules, while they are limited by predictive performance~\citep{xu2022subgraph}. Recently, there has been a surge in inductive KGC methods based on Graph Neural Networks (GNNs), inspired by the ability of GNNs to aggregate local information. For instance, GraIL~\citep{teru2020inductive} models enclose subgraphs of the target link to capture local topological structure, thereby possessing the inductive ability of emerging entities.
Motivated by GraIL and the message passing mechanism of GNN, some works~\citep{mai2021communicative,chen2021topology} have further utilized the enclosing subgraph structure and designed effective propagation strategies to model informative neighbors for inductive KGC. To explicitly enhance the prediction ability of unseen entities through semantic knowledge in the KGs, SNRI~\citep{xu2022subgraph} proposes a relational paths module to improve the performance of inductive KGC, while RMPI~\citep{geng2023relational} designs a novel relational message-passing network for fully inductive KGC with both unseen entities and relations. Despite the promising results yielded by these models, they are limited to information redundancy in modeling irrelevant entities and relations. AdaProp~\citep{zhang2023adaprop} is developed to learn an adaptive propagation path and filter out irrelevant entities, achieving powerful performance. However, these models ignore the unconvincing knowledge (e.g., semantic inconsistencies of similar relations in context and inherent noise within KGs).
% inconsistency caused by different semantic expressions of similar relations and cannot consider the inherent noisy interactions in the KGs, thus limiting their ability to model emerging entities. 
% The presence of semantic inconsistency and noisy associations within the KGs impairs the ability to reason on subgraphs, reducing the prediction accuracy of emergent entities.
For example, the same semantic ``\textit{the location is}'' with different relations \textit{located\_in} and \textit{lie\_in} may lead to inconsistent knowledge expressions in the context ``(Alibaba, \textit{lie\_in}, Hangzhou), and (Hangzhou, \textit{located\_in}, China)'', increasing the complexity of subgraph reasoning, which deduces the prediction performance on inductive situations. Additionally, logical reasoning may be misled to conclude a confused fact (Obama, \textit{live\_in}, New York) from the reasoning chain (Obama, \textit{work\_in}, The White House, \textit{located\_in}, New York) due to the presence of the false positive fact: The White House is \textit{located in} New York. More cases are shown in Appendix A.4.

% \begin{figure}[t]
% \centering
% \includegraphics[width=1\columnwidth]{figs/S$^2$DN_intro_noise_1.pdf} % Reduce the figure size so that it is slightly narrower than the column. Don't use precise values for figure width. This setup will avoid overfull boxes.
% \caption{\textbf{Left}: S$^2$DN outperforms GraIL in terms of ranking Hits@10 on contaminated KGs with different noise ratios (i.e., \textit{high robustness}). \textbf{Right}: S$^2$DN shows a higher percentage of the relation \textit{edited\_by} being converted to other relationships (such as \textit{written\_by}) with similar semantics (i.e., \textit{high semantic consistency}).} 
% \label{fig:intro_noise}
% \end{figure}

Motivated by the above observations, the inductive KGC presents two main challenges: (i) \textit{\textbf{Inconsistency}}: inconsistent representations of relations with the same semantics in context, and (ii) \textit{\textbf{Noisy}}: the presence of inevitable noise in KGs that is difficult to ignore.
In response to these challenges, we introduce S$^2$DN, a semantic structure-aware denoising network designed to maintain consistent semantics and filter out noisy interactions, thereby enhancing robustness in inductive KGC. 
Drawing inspiration from the successful application of smoothing technologies in image denoising~\citep{ma2018deep,guo2019smooth}, we have developed a semantic smoothing module to generalize similar relations with blurred semantics. Additionally, to eliminate task-irrelevant noise and provide supplementary knowledge, we introduce a structure refining module to retain reliable interactions in a learnable manner. By integrating general semantics and reliable structure, S$^2$DN denoise unconvincing knowledge from a semantic-structure synergy perspective.
As depicted in Figure~\ref{fig:intro_noise}, S$^2$DN demonstrates superior inductive prediction performance (Hits@10) compared to GraIL across different KGs under various noise levels. Additionally, S$^2$DN ensures improved semantic consistency, as evidenced by a high percentage of the relation \textit{edited\_by} being smoothed to others (e.g., \textit{written\_by} and \textit{produced\_by}) with similar semantics.

The contributions of S$^2$DN are summarized: 1) We address inductive KGC from a novel perspective by adaptively reducing the negative impact of semantic inconsistency and noisy interactions; 2) We innovatively propose a semantic smoothing module to generalize KG relations dynamically by blurring similar relations into consistent knowledge semantics; 3) To emphasize reliable interactions, we introduce the structure refining module to adaptively filter out noise and offer additional knowledge. 4) Extensive experiments on benchmark datasets and contaminated KGs demonstrate that S$^2$DN outperforms the baselines in inductive KGC.
% \begin{itemize}[leftmargin=*]
%     \item We address inductive KGC from a novel perspective by adaptively reducing the negative impact of semantic inconsistency and noisy interactions. 
%     \item We innovatively propose a semantic smoothing module to generalize KG relations dynamically by blurring similar relations into consistent knowledge semantics.
%     % Inspired by the success of smooth technologies in image denoising, we innovatively propose a semantic smoothing module to generalize KG relations dynamically by blurring similar relations into consistent knowledge semantics.
%     \item To emphasize reliable interactions, we introduce the structure refining module to adaptively filter out noise and offer additional knowledge.
%     \item Extensive experiments on benchmark datasets and contaminated KGs demonstrate that S$^2$DN outperforms the state-of-the-art baselines in inductive KGC.
% \end{itemize}

%% file: rel_work.tex
\section{Related Works}
\noindent\textbf{Inductive Knowledge Graph Completion.}
% Inductive Knowledge Graph Completion (KGC) endeavors to tackle the inherent incompleteness of dynamic Knowledge Graphs (KGs) by identifying missing interactions among emerging entities~\citep{inductive_kgc}. 
Previous methods fall into two main categories: \textit{Rule-based} and \textit{GNN-based} approaches. 
%% rule-base method %%
The rule-based methods are independent of entities by mining logical rules for explicit reasoning. For instance, NeuralLP~\citep{yang2017differentiable} and DRUM~\citep{sadeghian2019drum} learn logical rules and their confidence simultaneously in an end-to-end differentiable manner. Similarly, IterE~\citep{zhang2019iteratively} and RNNLogic~\citep{qu2020rnnlogic} treat logic rules as a latent variable, concurrently training a rule generator alongside a reasoning predictor utilizing these rules. RLogic~\cite{cheng2022rlogic} enhances rule learning in knowledge graphs by defining a predicate representation learning-based scoring model and incorporating deductive reasoning through recursive atomic models, improving both effectiveness and computational efficiency. To learn high-quality and generalizable rules, NCRL~\cite{cheng2023neural} proposes to compose logical rules by detecting the best compositional structure of a rule body and breaking it into small compositions to infer the rule head.
%% gnn-base method %%
While rule-based methods have shown comparable prediction performance, they often overlook the surrounding structure of the target link, resulting in limited expressive ability in inductive scenarios~\citep{xu2022subgraph}. Recently, there has been a shift towards leveraging GNNs for KGC tasks~\citep{teru2020inductive,zhang2023adaprop}. For example, LAN~\citep{wang2019logic} learns the embeddings of unseen entities by aggregating information from neighboring nodes, albeit restricted to scenarios where unseen entities are surrounded by known entities. GraIL~\citep{teru2020inductive} and CoMPILE~\citep{mai2021communicative} address this limitation by modeling the enclosing subgraph structure around target facts. However, these approaches neglect the neighboring relations surrounding the target triple. Thus, SNRI~\citep{xu2022subgraph} and RED-GNN~\citep{zhang2022knowledge} fully exploit complete neighboring relations and paths within the enclosing subgraph. 
While LogCo~\citep{pan2022inductive} tackles the challenge of deficient supervision of logic by leveraging relational semantics. 
RMPI~\citep{geng2023relational} proposes a relational message-passing network to utilize relation patterns for subgraph reasoning. These methods ignore the negative impact of task-irrelevant entities. AdaProp~\citep{zhang2023adaprop} designs learning-based sampling mechanisms to identify the semantic entities. 
Although these methods have achieved promising results, they suffer from (i) inconsistent semantics of similar relations, and (ii) inherent noisy associations within KGs. Our work proposes to smooth semantic relations and learn reliable structures to tackle these limitations.

\begin{figure} % 40%
  \centering % 
  \includegraphics[width=0.95\columnwidth]{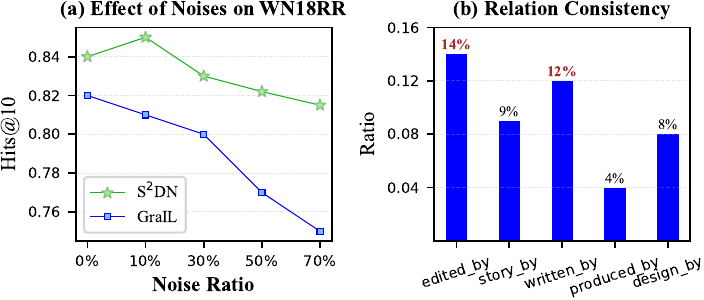} % 
  \caption{(a) S$^2$DN outperforms GraIL in terms of Hits@10 on noisy KGs with different noise ratios (i.e., \textit{high robustness}). (b) The relation \textit{edited\_by} shows a high percentage of being converted to other relations (enumerated on the x-axis) with similar semantics (i.e., \textit{high semantic consistency}).} %
  \label{fig:intro_noise}
\end{figure}

\noindent\textbf{Denoising Methods in Knowledge Graphs.}
%% traditional methods
% KGs have led to the proposal of numerous construction and application technologies. However, 
The presence of noise within KGs is a common issue stemming from the uncertainty inherent in learning-based construction methods~\citep{KG_quality,pujara2017sparsity}. 
%% learning-based methods
Denoising on KGs has been applied to the recommendation~\citep{fan2023graph} and social networks~\citep{quan2023robust}. For instance, ADT~\citep{wang2021denoising} and KRDN~\citep{zhu2023knowledge} designed a novel training strategy to prune noisy interactions and implicit feedback during training. RGCF~\citep{tian2022learning} and SGDL~\citep{gao2022self} proposed a self-supervised robust graph collaborative filtering model to denoise unreliable interactions and preserve the diversity for recommendations. However, these methods are limited in their ability to denoise noisy interactions in domain-specific networks (e.g., recommendation and social networks). 
% Moreover, they struggle to work well on the general KGC tasks. 
Some approaches~\citep{denoise_kg_entity,springer_denoise_rule} attempt to adopt rule-based triple confidence and structural entity attributes to capture noise-aware KG embedding. Despite achieving promising results, these methods overlook inconsistent semantic relations and work for transductive KGC reasoning. Inspired by the smoothing insight in image denoising~\citep{ma2018deep,guo2019smooth}, we propose a method to smooth the complex relations within KGs for inductive KGC. By doing so, we aim to eliminate inconsistent semantics and extract reliable structures in local subgraphs, particularly effective in inductive scenarios.

%% file: preliminary.tex
% \section{Notations}

\section{Preliminary}
% \noindent\textbf{\underline{Biomedical Knowledge Graph}.} 
\noindent\textbf{Knowledge Graphs.} KGs contain structured knowledge about real-world facts, including common concepts, entity attributes, or external commonsense.
We define a KG as a heterogeneous graph $\mathcal{G} = \{(h,r,t)|h,t\in \mathcal{E}, r\in \mathcal{R}\}$ where each triple $(h,r,t)$ describes a relation $r$ between the head entity $h$ and tail entity $t$ as a fact. 

\noindent\textbf{Enclosing Subgraph.}
Following GraIL~\citep{teru2020inductive}, when given a KG $\mathcal{G}$ and a triple $(u,r,v)$, we extract an enclosing subgraph $g=(V,E)$ surrounding the target triple. Initially, we obtain the $k$-hop neighboring nodes $\mathcal{N}_{k}(u)=\{s|d(u,s)\leq k\}$ and $\mathcal{N}_{k}(v)=\{s|d(v,s)\leq k\}$ for both $u$ and $v$, where $d(\cdot,\cdot)$ represents the shortest path distance between given nodes on $\mathcal{G}$. Subsequently, we obtain the nodal intersection $V=\{s|s\in \mathcal{N}_{k}(u)\cap \mathcal{N}_{k}(v)\}$ as vertices of the subgraph. Finally, we draw the triples linked by the set of nodes $V$ from $\mathcal{G}$ as $g=(V,E)$. 

% \noindent\textbf{Definitions}
\noindent\textbf{Problem Definition.} We concentrate on predicting missing links between emerging entities within a knowledge graph $\mathcal{G}$ (i.e., inductive KGC). This prediction is achieved by adaptively smoothing relational semantics and refining reliable structures. We define the problem of inductive KGC as a classification task, aiming to estimate the interaction probability of various relations inductively. Specifically, given an unknown fact $(u,r,v)$ where either $u$ or $v$ is an emerging entity, we propose a model to predict the interaction probability denoted as $p_{(u,r,v)} = \mathcal{F}((u,r,v)|\Theta, \mathcal{G}, g)$, where $\Theta$ is the trainable parameters of S$^2$DN.
% This paper focuses on predicting the missing links between emerging entities based on enclosing subgraph $g$ within KG $\mathcal{G}$ by adaptively smoothing relational semantics and refining reliable structure. We consider the inductive knowledge graph completion a classification task. Our goal is to estimate the interaction probability of various relations inductively. For a given unknown fact $(u,r,v)$, in which $u$ or $v$ is emerging, we propose a model to predict the interaction probability denoted as $p_{(u,r,v)} = \mathcal{F}((u,r,v)|\Theta, \mathcal{G}, g)$.

% \noindent\textbf{Definition of Smoothing.} \textit{Given the relations $R$ within the enclosing subgraph $g$, }

%% file: method.tex
% \section{The S$^2$DN Framework}
% \begin{figure}[t]
% \centering
% \includegraphics[width=1\columnwidth]{figs/robust_overview.pdf} % Reduce the figure size so that it is slightly narrower than the column. Don't use precise values for figure width. This setup will avoid overfull boxes.
% \caption{The S$^2$DN framework comprises two modules for robust predicting links in a given KG inductively: (1) Smoothing relational semantics by blurring similar relations adaptively; (2) Refining the structure of subgraphs by learning reliable interactions dynamically.}
% \label{fig:overview}
% \end{figure}

\section{Proposed Method}
\subsection{The Framework of S$^2$DN}
\begin{figure} % 40%
  \centering % 
  \includegraphics[width=0.9\columnwidth]{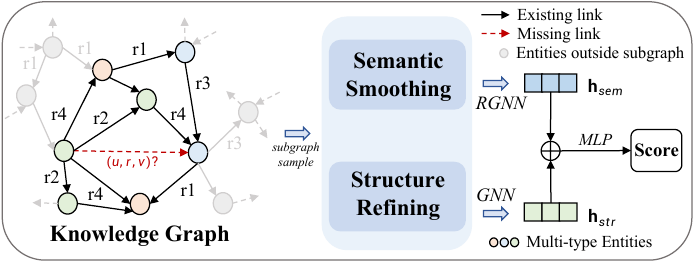} % 
  \caption{The S$^2$DN framework comprises two modules for inductively predicting links in a given KG : (1) Smoothing relational semantics by blurring similar relations adaptively; (2) Refining the structure of subgraphs by learning reliable interactions dynamically.}
\label{fig:overview}
\end{figure}

\noindent\textbf{Overview.}
S$^2$DN reasons on the enclosing subgraph surrounding the target link inductively from both semantic and structure perspectives, as illustrated in Figure~\ref{fig:overview}. To identify the unknown link $(u,r,v)$, S$^2$DN incorporates two key components: \textit{Semantic Smoothing} and \textit{Structure Refining}. Semantic Smoothing adaptively merges relations with similar semantics into a unified representation, ensuring consistency in representation space. In parallel, Structure Refining dynamically focuses on filtering out task-irrelevant facts surrounding the target link and incorporates additional knowledge, thus improving the reliability of interactions. The refining process works in tandem with the previously smoothed relations to predict unknown links involving new entities more effectively. After obtaining the smoothed and refined subgraphs, we model them using a Relational Graph Neural Network (RGNN,~\citep{schlichtkrull2018modeling}) and a Graph Neural Network (GNN,~\citep{kipf2016semi}), respectively. Finally, the embeddings of smoothed and refined subgraphs are concatenated and fed into a classifier to predict the interaction probability of the target link $(u,r,v)$.

% S$^2$DN is a semantic-aware denoising network that can be used for robust inductive KGC. As illustrated in Figure~\ref{fig:overview}, for a given unknown link $(u,?,v)$, S$^2$DN reasons on the enclosing subgraph around the target link inductively. To predict the unknown link $(u,?,v)$, 
% S$^2$DN adopts two key components: (1) \textbf{Semantic Smoothing}, which adaptively blurs relations with the same semantics into unified representation in a learnable manner, keeping the semantics of relations consistent;
% and (2) \textbf{Structure Refining} focuses on reliable interactions and filters out task-irrelevant facts surrounding the target link dynamically, which corporates with smoothed relations for robust predicting unknown links on new entities. After obtaining the smoothed and refined subgraphs, we adopt a relational graph neural network (RGNN~\citep{schlichtkrull2018modeling}) and graph neural network (GNN~\citep{kipf2016semi}) to model them, respectively. Finally, we concat the embeddings of smoothed and refined subgraphs and feed them into the MLP to predict the interaction probability of the target link $(u,?,v)$.

\begin{figure} % 40%
  \centering % 
  \includegraphics[width=1\columnwidth]{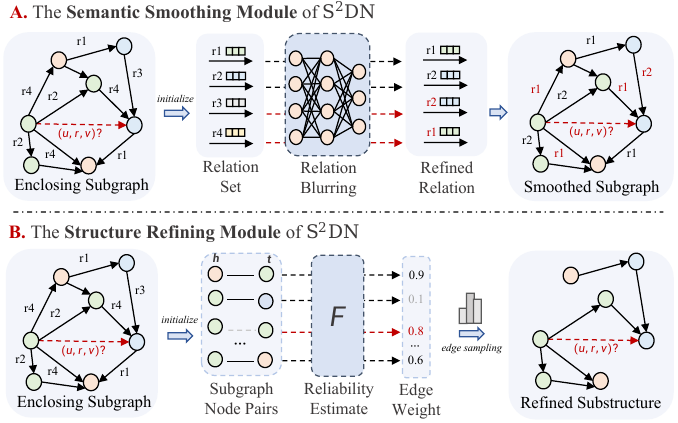} % 
  \caption{The architecture of \textbf{Semantic Smoothing} and \textbf{Structure Refining} modules of S$^2$DN.}
\label{fig:modules}
\end{figure}

% \noindent\textbf{Semantic Smoothing.}\label{sec:semantic}
\noindent\textbf{Semantic Smoothing.}\label{sec:semantic}
KGs often suffer from semantic inconsistencies in their relationships. For example, in the contexts (Alibaba, \textit{lie\_in}, Hangzhou) and (Hangzhou, \textit{located\_in}, China), the relations ``\textit{located\_in}'' and ``\textit{lie\_in}'' represent the same semantic meaning, ``\textit{the location is}''. These inconsistencies lead to discrepancies in the representation space~\citep{pujara2017sparsity,KG_quality}.
To mitigate this limitation, inspired by the pixel smoothing insight of image denoising~\citep{ma2018deep,guo2019smooth}, we propose a semantic smoothing module depicted in Figure~\ref{fig:modules}A. This module blurs similar relations while preserving the smoothed relational semantics, aiming to alleviate the negative impact of potential inconsistency. To achieve this, we first identify the subgraph $g=(V,E)$ that surrounds the missing link $(u,r,v)$. 
Then a trainable strategy is employed to smooth embeddings $\mathbf{E}\in \mathbb{R}^{|\mathcal{R}|\times dim}$ of similar relations into consistent representation space, where $|\mathcal{R}|$ denotes the count of original relations and $dim$ is the embedding size of relation embedding. Subsequently, we define the smooth operation as follows:
% We smooth the relations as follows:
\begin{equation}
\label{eq1}
\begin{gathered}
    % \tilde{\mathcal{R}} = \mathop{\mathrm{argmax}}\limits_{r\in \mathcal{R}}(\mathrm{softmax}(\mathbf{E}\otimes\mathbf{W}^{\mathrm{T}} +b)), 
    w = \mathrm{softmax}(\mathbf{E}\otimes\mathbf{W}^{\mathrm{T}} +b),\\
    \tilde{\mathcal{R}}=\frac{\exp((\log w+G)/\tau)}{\sum_{r\in \mathcal{R}}\exp((\log w_r+G_r)/\tau)},
\end{gathered}
\end{equation}
where $w$ denotes smoothing weights and $\tilde{\mathcal{R}}$ is the smoothed relations from original relations $\mathcal{R}$. $\mathbf{W}\in \mathbb{R}^{|\mathcal{R}|\times dim}$ is trainable parameters, $G$ is a noise sampled from a Gumbel distribution, and $\tau$ represents the temperature parameter. We adopt the Gumbel Softmax trick~\cite{jang2016categorical}, facilitating differentiable learning over discrete outputs $w$.
This operation learns to categorize relations with consistent semantics in the context of $g$ into the same relation index. We refine the embeddings of relations $\tilde{\mathcal{R}}$ as follows:
\begin{equation}
\label{eq1}
\begin{gathered}
    % \tilde{\mathbf{E}} = \text{\textbf{one-hot}}(\tilde{\mathcal{R}})\otimes\mathbf{E},
    \tilde{\mathbf{E}} = \tilde{\mathcal{R}}\otimes\mathbf{E},
\end{gathered}
\end{equation}
where $\tilde{\mathbf{E}}$ denotes the smoothed embeddings from original representation $\mathbf{E}$ and $\otimes$ represents the operation of matrix multiplication. This process enables similar relations to be mapped into consistent representation space guided by downstream tasks.
% The relation embedding $\mathbf{E}$ is initialized by Xavier initializer~\citep{glorot2010understanding}. 
To prevent the loss of information caused by excessive smoothing of relations and contain further consistencies, we incorporate a trade-off objective designed to preserve generic information during the optimization process.
After obtaining the smoothed relations, we refined the enclosing subgraph $g$ by the new relations $\tilde{\mathcal{R}}$. Then a $L$-layer RGNN~\citep{schlichtkrull2018modeling,xu2022subgraph} is introduced to capture the global semantic representation of the refined $g$. Specifically, the updating function of the nodes over the blurred relation embedding $ \tilde{\mathbf{E}}$ in $l$-th layer is defined as:
\begin{equation}
\label{eq2}
    \begin{gathered}\mathbf{x}_i^l=\sum_{r\in \tilde{R}}\sum_{j\in\mathcal{N}_r(i)}\alpha_{i,r}\mathbf{W}_r^l\phi(\tilde{\mathbf{e}}_r^{l-1},\mathbf{x}_j^{l-1}),\\
    \alpha_{i,r}=\mathrm{sigmoid}\left(\mathbf{W}_1\left[\mathbf{x}_i^{l-1}\oplus\mathbf{x}_j^{l-1}\oplus \tilde{\mathbf{e}}_r^{l-1}\right]\right),
    \end{gathered}
\end{equation}
where $\mathcal{N}_r(i)$ and $\alpha_{i,r}$ denote the neighbors and the weight of node $i$ under the relation $r$, respectively. $\oplus$ indicates the concatenation operation. $\mathbf{W}_r^l$ represents the transformation matrix of relation $r$, $\mathbf{x}_i$  stands for the embedding of node $i$, $\tilde{\mathbf{e}}_r$ denotes the smoothed embedding under relation $r$, and $\phi$ is the aggregation operation to fuse the hidden features of nodes and relations. In addition, we initialize the entity (i.e., node) embedding $\mathbf{x}_i^0$ with the designed node features~\citep{teru2020inductive} and the original relation embedding $\mathbf{E}$ is initialized by Xavier initializer~\citep{glorot2010understanding}. The details of designed node features refer to Appendix B.1.3.
Finally, we obtain the global representation $\mathbf{h}_{sem}$ of the smoothed subgraph $g$ as follows:
\begin{equation}
\label{eq3}
\mathbf{h}_{sem}=\frac1{|V|}\sum_{i\in V}^{V}\sigma(f(\mathbf{x}_i^L)), 
\end{equation}
where $V$ is the node set of smoothed subgraph $g$ and $f(\cdot)$ denotes the feature transformation function. $\sigma$ indicates the activation function ReLU~\citep{nair2010rectified}.

\noindent\textbf{Structure Refining.}\label{sec:struc_refine}
To improve the precise estimation of noisy interactions within KGs, we propose a structure refining module specifically designed for the local enclosing subgraph, as depicted in Figure~\ref{fig:modules}B. This module dynamically adapts the reliable subgraph structure based on both node features and feedback from downstream tasks. The underlying concept is that nodes with similar features or structures are more prone to interact with each other compared to those with dissimilar attributes~\citep{zhang2020gnnguard,li2024gslb}. Our objective is to assign weights to all edges connecting the nodes using a reliability estimation function denoted as $F(\cdot,\cdot)$, which relies on the learned node features.
% This module is designed to dynamically adjust the reliable subgraph structure based on the node features and feedback from downstream tasks. The core idea is that nodes sharing similar features or structures are more likely to engage in interactions compared to those with dissimilar features or structures~\citep{zhang2020gnnguard}. Our goal is to assign weights to all edges connecting the nodes using a reliability estimation function denoted as $F(\cdot,\cdot)$, which is dependent on the learned node features.
% To achieve this, we aim to weighted all edges between the nodes set using reliability estimation function $F(\cdot,\cdot)$ based on the pretrained node features. 
% Then, the refined local subgraph can be generated by filtering out noisy edges with low weight and retaining the reliable links with larger ones.
Following this, the refined local subgraph is generated by removing low-weight noisy edges while retaining the more significant and reliable connections. To elaborate, when presented with an extracted enclosing subgraph $g=(V,E)$ surrounding the missing link $(u,r,v)$, we prioritize the degree of interaction over relations, thereby enriching the structural information of semantic smoothing modules. We conceptualize all potential edges between nodes as a collection of mutually independent Bernoulli random variables, parameterized by the learned attention weights $\pi$.
% Subsequently, the refined local subgraph is created by filtering out low-weight noisy edges and preserving the more substantial reliable links. Specifically, given an extracted enclosing subgraph $g=(V, E)$ around the missing link $(u, ?, v)$, we focus on the degree of interaction and ignore the relations to enhance the structural information of semantic smoothing modules. We model all possible edges between the nodes as a set of mutually independent Bernoulli random variables parameterized by the learned attention weights $\pi$.
\begin{equation}
\label{eq4}
    \tilde{g}=\bigcup_{i,j\in V}\left\{(i,j)\sim\mathrm{Ber}\left(\pi_{i,j}\right)\right\}.
\end{equation}
In this context, $V$ denotes the set of nodes within the enclosing subgraph, and $(i,j)\in E$ denotes the edge connecting nodes $i$ and $j$. We optimize the reliability probability $\pi$ concurrently with the downstream inductive KGC task. The value of $\pi_{i,j}$ characterizes the task-specific reliability of the edge $(i,j)$ where a smaller $\pi_{i,j}$ suggests that the edge $(i,j)$ is more likely to be noisy and thus should be assigned a lower weight or removed altogether. The reliable probability $\pi_{i,j} = F(i,j)$ for each edge between node pair $(i,j)$ can be calculated as follows:
\begin{equation}\label{eq5}
    \begin{aligned}
    &\pi_{i,j} =\mathrm{sigmoid}\left(Z(i)Z(j)^\mathrm{T}\right),
    \\
    &Z(i) =\mathbf{MLP}\left(\mathbf{X}\left(i\right)\right),
\end{aligned}
\end{equation}
where $\mathbf{X}\left(i\right)$ represents the feature of node $i$, $Z(i)$ is the learned embedding of node feature $\mathbf{X}\left(i\right)$, and $\mathbf{MLP}\left(\cdot\right)$ denotes a two-layer perceptron in this work (More detailed choices of $F(\cdot,\cdot)$ are discussed in Appendix C.4.4). Since the extracted enclosing subgraph $g$ is not differentiable when the probability $\pi$ is modeled as a Bernoulli distribution, we employ the reparameterization trick. This allows us to relax the binary entries $\mathrm{Ber}(\pi_{i,j})$ for updating the $\pi$:
\begin{equation}\label{eq6}
    \small
    \mathrm{Ber}(\pi_{i,j})\approx\mathrm{sigmoid}\left(\frac1t\left(\log\frac{\pi_{i,j}}{1-\pi_{i,j}}+\log\frac\epsilon{1-\epsilon}\right)\right),
\end{equation}
where $\epsilon\sim$ \textit{Uniform}$(0,1)$, $t\in \mathbb{R}^+$ indicates the temperature for the concrete distribution. With $t\geq 0$, the function undergoes smoothing with a well-defined gradient $\frac{\partial\mathrm{Ber}(\pi_{i,j})}{\partial \pi_{i,j}}$, which facilitates the optimization of learnable subgraph structure throughout the training process. After post-concrete relaxation, the subgraph structure becomes a weighted fully connected graph, which is computationally intensive. To address this, edges with a probability of less than 0.5 are removed from the subgraph, yielding the refined graph $\tilde{g}$. Following this refinement, $L$-layer GCNs~\citep{kipf2016semi} are applied to the refined subgraph using the designed node features (see Appendix B.1.3) to derive its global representation $\mathbf{h}_{str}$ as follows:
% The subgraph structure after the concrete relaxation is a weighted fully connected graph, which is computationally expensive. We hence drop the edges of the subgraph with a probability of less than 0.5 and get the refined graph $g^{'}$. Subsequently, we perform the $L$-layer GCNs~\citep{kipf2016semi} on the refined subgraph with the designed node features (see Section~\ref{node_feat}) to obtain its global representation $\mathbf{h}_{str}$ as follows:
\begin{equation}\label{eq7}
\begin{aligned}
        &h^l =\mathbf{GCN}\left(h^{l-1}, \tilde{g}\right),
         \\
        &\mathbf{h}_{str}=\frac1{|V|}\sum_{i\in V}^{V}\sigma(f(h^L(i))),
\end{aligned}
\end{equation}
where $\sigma(\cdot)$ represents the sigmoid activation function. $f(\cdot)$ is a multi-layer perceptron that denotes the feature transformation operation.

% \begin{figure}[t]
% \centering
% \includegraphics[width=1\columnwidth]{figs/robust_modules.pdf} % Reduce the figure size so that it is slightly narrower than the column. Don't use precise values for figure width. This setup will avoid overfull boxes.
% \caption{The architecture of \textbf{Semantic Smoothing} and \textbf{Structure Refined} modules of S$^2$DN.}
% \label{fig:modules}
% \end{figure}

\subsection{Theoretical Discussion of Smoothing}\label{sec:dis_smooth} \label{sec:discussion}
The smoothing operation can be used as a way to minimize the information bottleneck between the original semantic relations and the downstream supervised signals (i.e., labels). Following the standard practice in the method~\citep{tishby2000information}, given original relation embedding $\mathbf{E}$, smoothed relation embedding $\tilde{\mathbf{E}}$, and target $Y$, they follow the Markov Chain $<Y\rightarrow\mathbf{E}\rightarrow\tilde{\mathbf{E}}>$.
% $(\mathbf{E},\tilde{\mathbf{E}},Y)$ follow the Markov Chain $<Y\rightarrow\mathbf{E}\rightarrow\tilde{\mathbf{E}}>$.
% Large facts and knowledge in KG provide massive semantics. Specifically, entity types represent the description of objects in the real world, and relations represent the interaction between different objects. However, the definitions of relations are hand-drafted and highly impacted by subjective factors, and the relation semantics may be inconsistent under different contexts. For example, the different relations ``\textit{located\_in}'' and ``\textit{lie\_in}'' represent the same semantic ``\textit{the location of}'', which brings relational semantic inconsistency. This limits the predictive performance of inductive KGC methods. 

\noindent\textbf{Definition 1 (Information Bottleneck).}  \textit{For the input relation embedding $\mathbf{E}$ and the label of downstream task $Y$, the \textbf{Information Bottleneck} principle aims to learn the minimal sufficient representation $\tilde{\mathbf{E}}$:
}
\begin{equation}
    \tilde{\mathbf{E}}=\arg\min_{\tilde{\mathbf{E}}}-I(\tilde{\mathbf{E}};Y)+ I(\tilde{\mathbf{E}};\mathbf{E}),
\end{equation}
where $I(A;B)=H(A)-H(A|B)$ denotes the Shannon mutual information~\citep{cover1999elements}.
Intuitively, the first term $-I(\tilde{\mathbf{E}};Y)$ is the reasoning objective, which is relevant to downstream tasks. The second term $I(\tilde{\mathbf{E}};\mathbf{E})$ encourages the task-independent information of the original relational semantic dropped. Suppose $\mathbf{E}_n\in\mathbb{R}$ is a task-irrelevant semantic noise in original subgraph $g$, the learning process of $\tilde{\mathbf{E}}$ follows the Markov Chain $<(Y,\mathbf{E}_n)\rightarrow\mathbf{E}\rightarrow\tilde{\mathbf{E}}>$. The smoothed relation embedding $\tilde{\mathbf{E}}$ only preserves the task-related information in the observed embedding $\mathbf{E}$.

% \subsubsection{Lemma 1 (Smoothing Objective).}
\noindent\textbf{Lemma 1 (Smoothing Objective).}
\textit{Given the original relation embedding $\mathbf{E}$ within the enclosing subgraph $g$ and the label $Y\in\mathbb{Y}$, let $\mathbf{E}_n\in\mathbb{R}$ be a task independent semantic noise for $Y$. Denote $\tilde{\mathbf{E}}$ as the smoothed relations learned from $\mathbf{E}$, then the following inequality holds:
}
\begin{equation}
    I(\tilde{\mathbf{E}};\mathbf{E}_n)\leq I(\tilde{\mathbf{E}};\mathbf{E})-I(\tilde{\mathbf{E}};Y).
    \label{noise_invariance}
\end{equation}
The detailed proof refers to Appendix A.1. Lemma 1 shows that optimizing the objective in Eq. (\ref{noise_invariance}) is equivalent to encouraging $\tilde{\mathbf{E}}$ to be more related to task-relevant information by minimizing the terms $I(\tilde{\mathbf{E}};\mathbf{E})$ and $-I(\tilde{\mathbf{E}};Y)$. Therefore, we introduce a Kullback–Leibler (KL) loss~\citep{sun2022graph} to minimize the difference between original and smoothed relation embeddings and adopt the cross-entropy loss to maximize the mutual information between the smoothed relations and downstream tasks. 
% The specific optimization procedure is shown in Section~\ref{sec:optimiz}.

\subsection{Training and Optimization}\label{sec:optimiz}
% \noindent\textbf{Training and Optimization.}
This section delves into the prediction and optimization details of S$^2$DN within the framework of the inductive KGC task. Here, we view the inductive KGC as a classification task. Given a predicted link $(u,r,v)$, the link probability $p_{(u,r,v)}$ for the given link is computed using representations from both semantic and structural perspectives as follows:
% In this section, we present the details of the prediction and optimization of S$^2$DN under the inductive KGC task. Specifically, we consider the inductive KGC task a classification task. For a predicted link $(u,r,v)$, we calculate the link probability $p_{(u,r,v)}$ of the given link by using the representations from semantic and structure views as follows:
\begin{equation}\label{eq10}
p_{(u,r,v)}=\sigma(\mathbf{MLP}([\mathbf{h}_{sem}\oplus\mathbf{h}_{str}])),
\end{equation}
where $\oplus$ denotes the concatenate operation, $\mathbf{MLP}(\cdot)$ indicates a classifier here and $\sigma(\cdot)$ is the sigmoid activate function. Subsequently, we adopt the cross-entropy loss and introduce an objective to balance the difference between smoothed and original relations:
\begin{equation}\label{eq11} 
    \ell=-\sum_{r\in\mathcal{R}}\log(p_{(u,r,v)})y_{(u,r,v)}+\mathcal{D}(\tilde{\mathbf{E}}||\mathbf{E})+\lambda \Vert\Theta\Vert_2,
\end{equation}
where $y_{(u,r,v)}$ is the real label of the given link, $\Theta$ represents the trainable parameters of S$^2$DN, $\mathcal{D}$ denotes the KL loss, and $\lambda$ is a hyperparameter denoting the coefficient of the regular term. $\tilde{\mathbf{E}}$ and $\mathbf{E}$ denote the representations before and after relation smoothing in the subgraph of the current sample, respectively.

%% file: experiment.tex
\section{Experiments}
We carefully consider the following \textit{key} research questions:
% \begin{itemize}[leftmargin=*]
%     \item \textbf{RQ1:} Does S$^2$DN outperforms other state-of-the-art inductive KGC baselines?
%     \item \textbf{RQ2:} Are the proposed \textit{Semantic Smoothing} and \textit{Structure Refining} modules effective?
%     \item \textbf{RQ3:} Can S$^2$DN enhance the semantic consistency of the relations within the KG?
%     \item \textbf{RQ4:} Can S$^2$DN refine reliable substructure surrounding the target facts for downstream tasks?
% \end{itemize}
\textbf{RQ1)} Does S$^2$DN outperform other state-of-the-art inductive KGC baselines? \textbf{RQ2)} Are the proposed \textit{Semantic Smoothing} and \textit{Structure Refining} modules effective? \textbf{RQ3)} Can S$^2$DN enhance the semantic consistency of the relations and refine reliable substructure surrounding the target facts?

% \subsection{Datasets} 
\subsection{Experiment Setup}
We show more \textbf{\underline{details of the implementations}} of S$^2$DN and baselines in Appendix \textbf{B.1} and \textbf{B.2}.

\noindent\textbf{Dataset \& Evaluation.}
We utilize three widely-used datasets: WN18RR~\citep{dettmers2018convolutional}, FB15k-237~\citep{toutanova2015representing}, and NELL-995~\citep{xiong2017deeppath}, to evaluate the performance of S$^2$DN and baseline models. Following~\citep{teru2020inductive,zhang2023adaprop}, we use the same four subsets with increasing size of the three datasets. Each subset comprises distinct training and test sets (Appendix B.1.4). We measure the filtered ranking metrics \textbf{Hits@1}, \textbf{Hits@10}, and mean reciprocal rank (\textbf{MRR}), where larger values indicate better performance (Appendix B.1.1).

\noindent\textbf{Baselines.} We compare S$^2$DN against the rule- and GNN-based methods. The rule-based methods  are \textbf{NeuralLP}~\citep{yang2017differentiable}, \textbf{DRUM}~\citep{sadeghian2019drum}, and \textbf{A$^*$Net}~\cite{zhu2024net}.
% \textbf{RuleN}~\citep{meilicke2018fine}. 
The GNN-based models are  \textbf{CoMPILE}~\citep{mai2021communicative},  TAGT~\cite{chen2021topology}, \textbf{SNRI}~\citep{xu2022subgraph}, and \textbf{RMPI}~\citep{geng2023relational}. We show more details of some missing baselines in Appendix B.2.3.
Furthermore, we design two variants of S$^2$DN to verify the effectiveness of each module by removing: (i) the Semantic Smoothing module (called \textbf{S$^2$DN w/o SS}), (ii) the Structure Refining module (called \textbf{S$^2$DN w/o SR}).

\begin{table*}[t]
\centering
\resizebox{1\textwidth}{!}{%
\begin{tabular}{l|l|lll|lll|lll|lll} \toprule
\multicolumn{1}{c|}{} & \multicolumn{1}{c|}{\textbf{Avg.}}                                  & \multicolumn{3}{c}{\textbf{V1}}                                                            & \multicolumn{3}{c}{\textbf{V2}}                                                            & \multicolumn{3}{c}{\textbf{V3}}                                                             & \multicolumn{3}{c}{\textbf{V4}}                                                             \\ \cline{3-14} 
\multicolumn{1}{c|}{\multirow{-2}{*}{\textbf{Methods}}} & Hits@10 & {Hits@1} & {Hits@10} & MRR           & {Hits@1} & {Hits@10} & MRR           & {Hits@1} & {Hits@10} & MRR            & {Hits@1} & {Hits@10} & MRR        \\ \midrule
DRUM       & 64.15  & 57.92          & 74.37          & 64.27          & 54.71          & 68.93          & 59.46          & 33.98          & 46.18          & 37.63          & 55.66          & 67.13          & 60.11          \\ 
NeuralLP   & 64.15  & 55.32          & 74.35          & 62.04          & 51.99          & 68.91          & 57.26          & 29.96          & 46.23          & 36.13          & 55.61          & 67.13          & 59.19          \\
A$^*$Net    & 72.06    & 70.51          & 80.58          & 74.68          & 70.99          & 79.11          & 74.45          & 46.47          & 54.87          & 49.68          & 63.92          & 73.67          & 66.24          \\ \midrule
CoMPILE   & 74.43   & 74.20           & 82.97          & 78.59         & \underline{78.11}          & 79.84          & 79.01          & 50.33          & 59.75          & 54.44          & 72.71          & 75.17          & 72.71          \\
TAGT     & 73.28    & 69.15          & 82.45          & 75.45          & 75.42          & 78.68          & 77.43          & 50.08          & 58.60           & 54.29          & 71.97          & 73.41          & 73.22          \\
SNRI    & \underline{79.19}     & 70.47          & \underline{84.84}          & 76.31          & 72.68          & 82.09          & 77.04          & 50.99          & 67.52          & \underline{57.46}          & \underline{74.28}          & \textbf{82.33}          & \underline{77.76}          \\

RMPI    & 73.34     & \textbf{75.53} & 82.45          & \underline{79.43}          & 75.85          & 78.68          & 77.64          & 52.64          & 58.84          & 55.97          & 71.48          & 73.41          & 72.98          \\ \midrule
S$^2$DN & \textbf{81.23} & \underline{74.73}          & \textbf{87.64} & \textbf{79.89} & \textbf{78.23} & \textbf{85.60} & \textbf{81.16} & \textbf{52.89} & \textbf{69.52} & \textbf{58.10} & \textbf{75.33} & \underline{82.15} & \textbf{78.04} \\
S$^2$DN w/o SS     & 78.49          & 74.46                               & 84.11                                & 78.02                            & 77.21                               & \underline{83.01}                                & {78.87}                            & 49.92                               & \underline{68.34}                                & 57.11                            & 74.21                               & 78.53                       & 76.01                   \\
S$^2$DN w/o SR      & 76.31         & 73.31                               & 84.04                                & 77.61                            & 77.14                               & 81.63                                & \underline{80.28}                            & \underline{51.24}                               & 63.22                                & 55.75                            & 73.93                               & 76.34                       & 75.58   \\               

\bottomrule
\end{tabular}
}%
\caption{The performance (i.e., \textbf{Hits@1}, \textbf{Hits@10}, \textbf{MRR}, in percentage) of S$^2$DN on WN18RR dataset. The boldface denotes the highest score and the underline indicates the second-best one.}
\label{tab:wn18}
\end{table*}

\begin{table*}[t]
\centering
\resizebox{1\textwidth}{!}{%
\begin{tabular}{l|l|lll|lll|lll|lll} \toprule
\multicolumn{1}{c|}{} & \multicolumn{1}{c|}{\textbf{Avg.}}                                  & \multicolumn{3}{c}{\textbf{V1}}                                                            & \multicolumn{3}{c}{\textbf{V2}}                                                            & \multicolumn{3}{c}{\textbf{V3}}                                                             & \multicolumn{3}{c}{\textbf{V4}}                                                             \\ \cline{3-14} 
\multicolumn{1}{c|}{\multirow{-2}{*}{\textbf{Methods}}} & Hits@10 & {Hits@1} & {Hits@10} & MRR           & {Hits@1} & {Hits@10} & MRR           & {Hits@1} & {Hits@10} & MRR            & {Hits@1} & {Hits@10} & MRR        \\ \midrule
DRUM                                            & 55.11        & 30.47                               & 52.92                                & 39.33         & 27.84                               & 58.73                                & 39.95         & 26.08                               & 52.90                                 & 47.02          & 27.09                               & 55.88                                & 38.41          \\
NeuralLP                                            & 55.16    & 31.43                               & 52.92                                & 38.25         & 28.88                               & 58.94                                & 38.39         & 26.09                               & 52.90                                 & 37.40           & 25.89                               & 55.88                                & 35.96         \\
A$^*$Net                                               & 76.57    & 37.09                               & 57.66                                & 43.63         & 50.47                               & 77.84                                & 60.33         & \textbf{61.45}                               & \textbf{85.32}                                & 60.39          & 56.38                               & 85.46                                 & \underline{68.29}       \\ \midrule
CoMPILE                                             & 78.95    & 41.70                                & 63.17                                & 49.92         & 54.54                               & 81.07                                & 64.14         & 55.63                               & 84.45                                & \textbf{67.18}          & 54.77                               & 87.13                                & 67.89        \\
TAGT                                                 & 77.61   & 38.05                               & 63.41                                & 48.20          & 50.84                               & 81.07                                & 61.48         & 51.21                               & 80.87                                & 61.83          & 49.89                               & 85.08                                & 62.16         \\

SNRI                                            & \underline{79.87}        & 30.98                               & \underline{64.29}                                & 42.81         & 50.52                               & \underline{81.37}                                & \textbf{66.58}         & 53.29                               & \underline{84.87}                                & 64.70           & 54.21                               & 88.97                                & 62.01         \\

RMPI                                          & 78.00          & \underline{41.93}                      & 63.41                                & \underline{50.57}         & \underline{54.92}                               & 80.54                                & 64.46         & 53.87                               & 81.33                                & 63.67          & 52.91                               & 86.73                                & 64.31         \\ \midrule
S$^2$DN                                   & \textbf{81.25}         & \textbf{43.68}                               & \textbf{67.34}                       & \textbf{52.10} & \textbf{55.45}                      & \textbf{82.38}                       & \underline{64.80} & \underline{56.31}                      & 83.97                       & \underline{65.07} & \textbf{60.96}                      & \textbf{91.31}                       & \textbf{68.44} \\
S$^2$DN w/o SS                        & 79.68            & 40.48                                                            & 67.07                                                             & 48.97                            & 52.41                                                            & 80.96                                                             & 63.13                            & 53.34                                                            & 80.91                                                             & 63.66                            & \underline{57.21}                                                            & \underline{89.77}                                                    & 66.12                  \\
S$^2$DN w/o SR                     & 78.97               & 39.76                                                            & 64.15                                                             & 48.58                            & 47.59                                                            & 78.35                                                             & 58.74                            & 54.16                                                            & 85.14                                                             & 64.94                            & 56.32                                                            & 88.23                                                    & 67.01     
\\ \bottomrule
\end{tabular}
}%
\caption{The performance (i.e., \textbf{Hits@1}, \textbf{Hits@10}, \textbf{MRR}, in percentage) of S$^2$DN on FB15k-237 dataset. We mark the best score with bold font and the second best with underline.}
\label{tab:fb237}
\end{table*}

\subsection{Comparison with Baselines (\textbf{RQ1})} \label{sec:comparision}
To address \textbf{RQ1},
we present the performance comparison of S$^2$DN and baseline models in predicting missing links for emerging entities as shown in Tables~\ref{tab:wn18} and \ref{tab:fb237} (Refer to Appendix C.1 for NELL-995). Our results demonstrate that S$^2$DN achieves comparable performance to the baseline models across all datasets.

Specifically, we make the following observations: (1) S$^2$DN shows improved average performance over rule-based inductive methods with Hits@10 metrics of 9.2\%, 4.7\%, and 6.8\% on WN18RR, FB15k-237, and NELL-995 respectively. 
Furthermore, GNN-based methods like CoMPILE and TAGT outperform various rule-based approaches in ranking tasks on most datasets, indicating the effectiveness of GNN-based methods in leveraging neighboring information and structures for inductive KGC. (2) SNRI, which integrates local semantic relations, outperforms CoMPILE and TAGT  on multiple datasets, highlighting the importance of utilizing local semantic relations within KGs for inductive KGC. (3) RMPI, through efficient message passing between relations to leverage relation patterns for KGC based on graph transformation and pruning, outperforms SNRI, emphasizing the significance of fully exploiting relational patterns and pruning links to enhance subgraph reasoning effectiveness. (4) S$^2$DN surpasses other GNN-based subgraph reasoning methods on most datasets, indicating that denoising unconvincing knowledge by promoting the consistency of relations and reliability of structures significantly enhances inductive KGC performance. In summary, S$^2$DN enhances the inductive reasoning capabilities of GNN-based models by effectively keeping relational semantics consistent and eliminating unreliable links, unlike previous GNN-based methods that overlook the impact of noise within KGs.

% \usepackage{multirow}

\subsection{Ablation Study (\textbf{RQ2})}
We undertake an ablation study across all datasets for the inductive KGC. The results are depicted in Table~\ref{tab:wn18} and Table~\ref{tab:fb237} (Refer to Appendix C.1 for NELL-995), confirming the effectiveness of each module.

\noindent\textbf{\underline{S$^2$DN w/o SS.}} After removing the semantic smoothing module, there is a notable performance decline across most datasets for inductive subgraph reasoning. This finding underscores the efficacy of maintaining relation consistency within the encompassing subgraph. Consequently, an informative enclosing subgraph featuring semantically consistent relations holds the potential to enhance S$^2$DN.

\noindent\textbf{\underline{S$^2$DN w/o SR.}} The exclusion of the structure refining module results in performance degradation across various datasets. This observation highlights the inadequacy of unreliable subgraph structures in conveying information effectively for the downstream KGC task, failing to mitigate the impact of noisy interactions. Conversely, a dependable structure or pristine subgraph enhances inductive reasoning capabilities by disregarding potential noise and preserving reliable interactions.

% In summary, the findings illustrated in Table~\ref{tab:wn18}, Table~\ref{tab:fb237}, and Table~\ref{tab:nell_995} underscore the effectiveness of the semantic smoothing and subgraph structure refining modules in maintaining semantic relation consistency and reducing the impact of unreliable interactions.

\subsection{Robustness of S$^2$DN (\textbf{RQ3})}\label{sec:robust_exp}
\begin{figure}[t]
% \begin{wrapfigure}{r}{0.4\textwidth}
\centering
\subfigure[WN18RR]{
\centering
    \includegraphics[width=0.47\columnwidth]{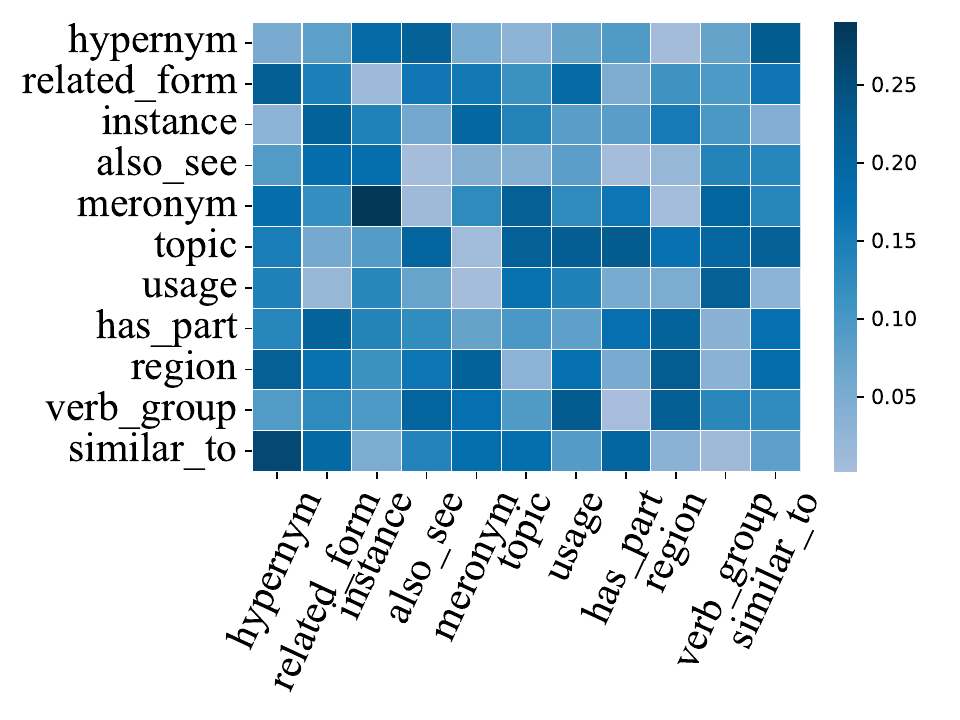} 
}
\subfigure[FB15k-237]{
    \centering
    \includegraphics[width=0.47\columnwidth]{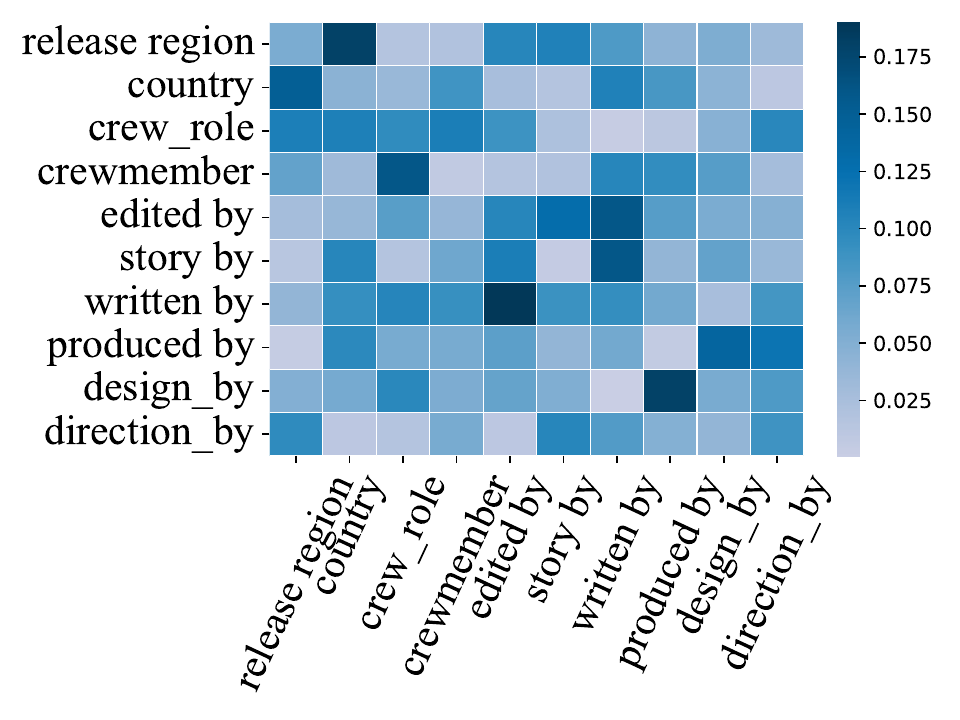} 
}
% \subfigure[NELL-995]{
%     \centering
%     \includegraphics[width=0.28\textwidth]{figs/nell995.pdf} 
% }
\caption{The transition ratio between the original and blurred relations on three datasets (V1 version). The element $m_{ij}$ in the matrix represents the proportion of the relation $i$ is smoothed to relation $j$.}
\label{fig:heatmap}
% \end{wrapfigure}
\end{figure}

\noindent\textbf{Semantic Consistency of Relations.} We conduct an experiment to analyze the semantic consistency. Specifically, as shown in Section~\ref{sec:semantic}, relations with similar semantics tend to be categorized into the same category, which benefits the semantic consistency of relations. During the inference process on the test dataset of WN18RR\_V1, FB15k-237\_V1, and NELL-995\_V1, we count the proportion $m_{ij}$ of relation $i$ that is categorized into another relation $j$ as the degree of semantic consistency. Thus, we can visualize the proportions $M=\{m_{ij}|0\leq i,j\leq|\mathcal{R}|\}$ of all relation pairs as shown in Figure~\ref{fig:heatmap}. For FB15k-237\_V1 and NELL-995\_V1, we singled out relationships related to the topic of \textit{movie} and \textit{sport}, respectively. We observe from Figure~\ref{fig:heatmap} that relations with similar semantics have a higher transformation rate than those with dissimilar semantics (e.g, the similar relations \textit{instance} and \textit{meronym} in Figure~\ref{fig:heatmap}a, \textit{edited by} and \textit{written by} in Figure~\ref{fig:heatmap}b, and \textit{athlete\_sport} and \textit{athlete\_team} in Figure~\ref{fig:heatmap}c). This phenomenon also proves that the Semantic Smoothing module can unify relations with similar semantics, thus maintaining the semantic consistency of the knowledge graph relations.

\begin{table}

\centering
\resizebox{1\columnwidth}{!}{%
\begin{tabular}{l|l|cccc} \toprule
\textbf{Noise Type}                 & \textbf{Model} & \multicolumn{1}{l}{\textbf{0\%}} & \textbf{15\%} & \textbf{35\%} & \textbf{50\%} \\ \midrule
\multirow{4}{*}{\textbf{Semantic}}  & RMPI                & $82.46$                          & $81.07_{\underline{ 1.7\%}}$           & $78.98_{\underline{ 4.2\%}}$         & $75.31_{\underline{ 8.7\%}}$         \\
                                    & S$^2$DN w/o SS          & $86.11$                          & $85.31_{\underline{ 0.9\%}}$         & $82.34_{\underline{ 4.4\%}}$         & $80.09_{\underline{ 7.0\%}}$         \\
                                    & S$^2$DN w/o SR          & $84.04$                          & $83.00_{\underline{ 1.2\%}}$         & $82.13_{\underline{ 2.3\%}}$         & $80.98_{\underline{ 3.6\%}}$         \\
                                    & S$^2$DN                 & $87.64$                          & $86.03_{\underline{ 1.8\%}}$         & $84.89_{\underline{ 3.1\%}}$         & $83.12_{\underline{ 5.2\%}}$         \\ \midrule
\multirow{4}{*}{\textbf{Structure}} & RMPI                & $82.46$                          & $80.79_{\underline{ 2.0\%}}$         & $78.31_{\underline{5.1\%}}$         & $76.24_{ \underline{7.5\%}}$         \\
                                    & S$^2$DN w/o SS          & $86.11$                          & $85.07_{\underline{ 1.2\%}}$         & $83.87_{\underline{ 2.6\%}}$         & $82.05_{\underline{ 4.7\%}}$         \\
                                    & S$^2$DN w/o SR          & $84.04$                          & $82.88_{\underline{ 1.4\%}}$         & $80.32_{\underline{ 4.4\%}}$         & $78.02_{\underline{ 7.2\%}}$         \\
                                    & S$^2$DN                 & $87.64$                          & $86.88_{\underline{ 0.9\%}}$         & $85.79_{\underline{ 2.1\%}}$         & $84.98_{\underline{ 3.0\%}}$   \\ \bottomrule     
\end{tabular}%
}
\caption{The results (\textbf{Hits@10}) of S$^2$DN on \textbf{WN18RR\_V1} under different noise ratios. The \underline{underline} indicates the drop rate of performance over noisy KGs.}
\label{tab:robust_wn}
\end{table}

\begin{table}

\centering
\resizebox{1\columnwidth}{!}{%
\begin{tabular}{l|l|cccc} \toprule
\textbf{Noise Type}                 & \textbf{Model} & \multicolumn{1}{l}{\textbf{0\%}} & \textbf{15\%} & \textbf{35\%} & \textbf{50\%} \\ \midrule
\multirow{4}{*}{\textbf{Semantic}}  & RMPI                & $63.41$                          & $61.76_{\underline{ 2.6\%}}$           & $59.01_{\underline{ 6.9\%}}$         & $57.29_{\underline{ 9.7\%}}$         \\
                                    & S$^2$DN w/o SS          & $67.07$                          & $65.22_{\underline{ 2.7\%}}$         & $63.03_{\underline{ 6.0\%}}$         & $61.89_{\underline{ 7.7\%}}$         \\
                                    & S$^2$DN w/o SR          & $64.15$                          & $63.01_{\underline{ 1.7\%}}$         & $62.33_{\underline{ 2.8\%}}$         & $60.14_{\underline{ 6.3\%}}$         \\
                                    & S$^2$DN                 & $67.34$                          & $66.05_{\underline{ 1.9\%}}$         & $65.78_{\underline{ 2.3\%}}$         & $64.97_{\underline{ 3.5\%}}$         \\ \midrule
\multirow{4}{*}{\textbf{Structure}} & RMPI                & $63.41$                          & $62.79_{\underline{ 0.9\%}}$         & $60.54_{\underline{ 4.5\%}}$         & $58.37_{\underline{ 7.9\%}}$         \\
                                    & S$^2$DN w/o SS          & $67.07$                          & $66.09_{\underline{ 1.4\%}}$         & $64.98_{\underline{ 3.1\%}}$         & $63.03_{\underline{ 6.0\%}}$         \\
                                    & S$^2$DN w/o SR          & $64.15$                          & $62.21_{\underline{ 3.0\%}}$         & $60.07_{\underline{ 6.3\%}}$         & $58.89_{\underline{ 8.1\%}}$         \\
                                    & S$^2$DN                 & $67.34$                          & $66.19_{\underline{ 1.7\%}}$         & $65.53_{\underline{ 2.6\%}}$         & $64.89_{\underline{ 3.6\%}}$   \\ \bottomrule     
\end{tabular}%
}
    \caption{The performance (\textbf{Hits@10}) of S$^2$DN on \textbf{FB15k-237\_V1} under various noise ratios. The \underline{underline} denotes the degree of performance decline over noisy KGs.}
\label{tab:robust_fb15k}
\end{table}
% \vspace{-1em}

\noindent\textbf{Reliability of S$^2$DN on Noisy KGs.}
We generate different proportions of \textit{semantic} and \textit{structural} negative interactions (i.e., 5\%, 15\%, 35\%, and 50\%) to contaminate the training KG. Semantic noises are created by randomly replacing the relations with others from known triples (e.g, the relation \textit{edited\_by} is replaced by \textit{written\_by}), while the structure noises are generated by sampling unknown triples from all \textit{entity-relation-entity} combinations. We then compare the performance of RMPI, S$^2$DN, and its variants on noisy KGs. As shown in Tables~\ref{tab:robust_wn} and \ref{tab:robust_fb15k} (Appendix C.2 for NELL-995), the performance of inductive KGC and their corresponding degradation ratio under different noise levels. 

As we introduce more noise, the performance of all methods declines under both semantic and structural settings. This decline is attributed to the dilution of expressive power caused by the randomly introduced false facts. Notably, S$^2$DN and its variants demonstrate less deduction compared to RMPI across most noisy KGs, suggesting that filtering out irrelevant interactions aids inductive reasoning on subgraphs. The variant S$^2$DN w/o SS exhibits superior performance on structurally noisy KGs compared to semantically noisy ones, highlighting the efficacy of the semantic smoothing module in modeling consistent relations. Conversely, the variant S$^2$DN w/o SR outperforms others on semantic noisy KGs rather than structural ones, indicating the importance of the structure refining module in uncovering informative interactions. Furthermore, S$^2$DN demonstrates comparable performance to other methods on both types of noisy KGs, indicating that S$^2$DN, equipped with semantic smoothing and structure refining modules, possesses enhanced subgraph reasoning capabilities.
% by emphasizing informative facts. 
This observation shows S$^2$DN can effectively mitigate unconvincing knowledge while providing reliable local structure.

\begin{figure}
\centering
\subfigure[GraIL, FB15k-237]{
\centering
    \includegraphics[width=0.4\columnwidth]{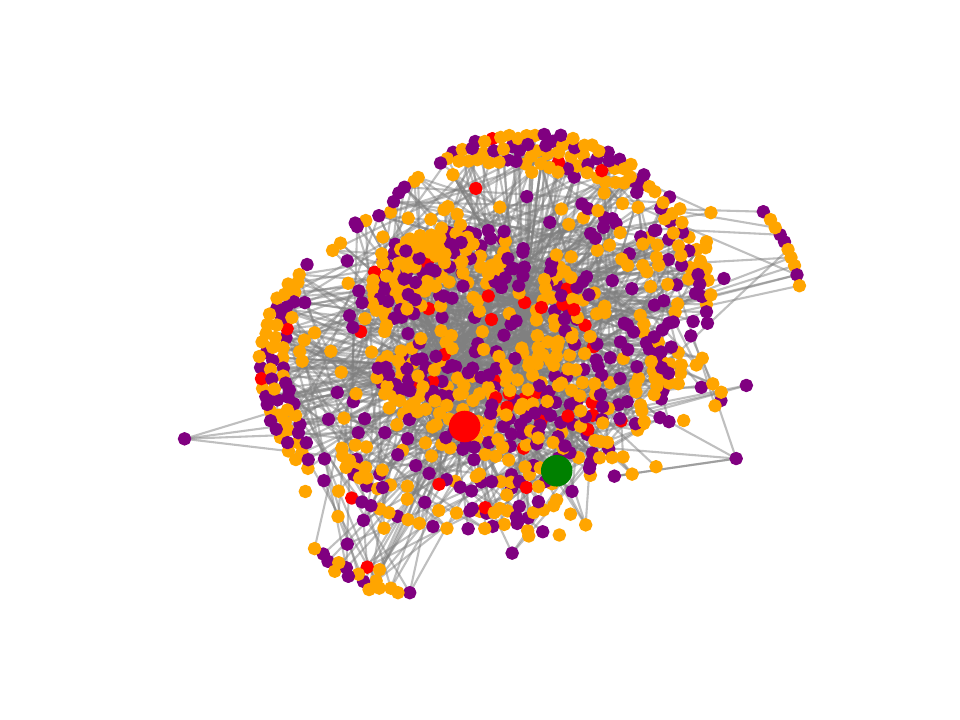}
}
\subfigure[S$^2$DN, FB15k-237]{
    \centering
    \includegraphics[width=0.4\columnwidth]{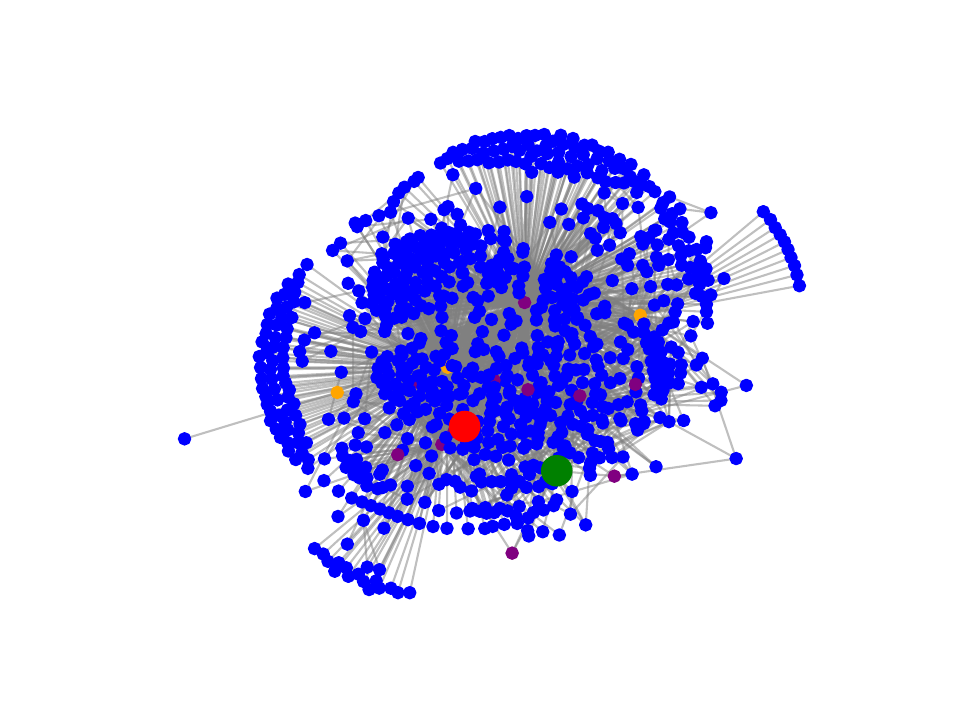} 
}

\caption{The big red and green nodes represent the source and target entities. The small nodes in red, orange, and blue are shared entities involved in $1-3$ hops between source and target nodes. The purple nodes indicate unshared entities.}
\label{fig:visualization}
% % \end{figure}
% \end{wrapfigure}
\end{figure}

\noindent\textbf{Visualization of Subgraphs.} To explicitly demonstrate the ability of S$^2$DN to provide reliable links to downstream tasks, we designed a case study on FB15k-237. We visualize the exemplar reasoning subgraph of S$^2$DN (i.e., the refined subgraph) and GraIL (i.e., the original subgraph) models for different queries (Appendix C.3 for more cases) in Figure~\ref{fig:visualization}. As illustrated in Figure~\ref{fig:visualization}, we observe that compared to GraIL, S$^2$DN can provide more knowledge for enhanced subgraph reasoning while retaining the original reliable information. For example, Figure~\ref{fig:visualization}(a) and Figure~\ref{fig:visualization}(b) show the subgraphs from GraIL and S$^2$DN have a similar layout, while S$^2$DN offers more links and filter out irrelevant interaction between source and target entities. This indicates that S$^2$DN is effective in subgraph reasoning inductively by a structure-refined mechanism.

%% file: appendix.tex
\section{Technical Appendix}
\section{A. S$^2$DN}
\subsection{A.1 Proof of Lemma 1 (\textit{Smoothing Objective})}\label{lemma_proof}
We provide the proof of Lemma 1.

\noindent\textit{Proof.} We prove Lemma 1 following the strategy of Proposition 3.1 in~\citep{achille2018emergence}. Suppose $\mathbf{E}$ is defined by $Y$ and $\mathbf{E}_n$, and $\tilde{\mathbf{E}}$ depends on $\mathbf{E}_n$ only through $\mathbf{E}$. We define the Markov Chain $<(Y,\mathbf{E}_n)\rightarrow\mathbf{E}\rightarrow\tilde{\mathbf{E}}>$. According to the data processing inequality (DPI), we have

\begin{equation}
    \begin{aligned}
        I(\tilde{\mathbf{E}};\mathbf{E})& \begin{aligned}\geq I(\tilde{\mathbf{E}};Y,\mathbf{E}_n)\end{aligned}  \\
        &\begin{aligned}=I(\tilde{\mathbf{E}};\mathbf{E}_n)+I(\tilde{\mathbf{E}};Y|\mathbf{E}_n)\end{aligned} \\
    &=I(\tilde{\mathbf{E}};\mathbf{E}_n)+H(Y|\mathbf{E}_n)-H(Y|\mathbf{E}_n;\tilde{\mathbf{E}}).
    \end{aligned}
\end{equation}
As we know, $\mathbf{E}_n$ is task-irrelevant noise independent of $Y$. Thus, we have $H(Y|\mathbf{E}_n)=H(Y)$ and $H(Y|\mathbf{E}_n;\tilde{\mathbf{E}})\leq H(Y|\tilde{\mathbf{E}})$. Then, we have 
\begin{equation}
    \begin{aligned}
I(\tilde{\mathbf{E}};\mathbf{E})& \begin{aligned}\geq I(\tilde{\mathbf{E}};\mathbf{E}_n)+H(Y|\mathbf{E}_n)-H(Y|\mathbf{E}_n;\tilde{\mathbf{E}})\end{aligned}  \\
&\begin{aligned}\geq I(\tilde{\mathbf{E}};\mathbf{E}_n)+H(Y)-H(Y|\tilde{\mathbf{E}})\end{aligned} \\
&=I(\tilde{\mathbf{E}};\mathbf{E}_n)+I(\tilde{\mathbf{E}};Y).
\end{aligned}
\end{equation}
Finally, we obtain $I(\tilde{\mathbf{E}};\mathbf{E}_n)\leq I(\tilde{\mathbf{E}};\mathbf{E})-I(\tilde{\mathbf{E}};Y).$

\subsection{A.2 Algorithm}
The full training process of S$^2$DN is shown in Algorithm~\ref{algo:full_process}. At the beginning, we initialize the entity embedding $\mathbf{X}$ and relation embedding $\mathbf{E}$ by designed features and Xavier initializer, respectively. Given a sample\footnote{We take one sample for easy understanding, a batch of samples are fed into S$^2$DN in practice.} $((u,r,v), y_{(u,r,v)})$ from training data $\mathbf{U}$, we extract its enclosing subgraph $g=(V,E)$ and feed $g$ into the semantic smoothing and structure refining modules. In the flow of the semantic smoothing module, we blur the relations of $g$ with consistent semantics into a unified embedding space and get the blurred relations $\tilde{R}$. The blurred relations $\tilde{R}$ are adopted to calculate the smoothed relational embedding $\tilde{\mathbf{E}}$ (see Eq. 1). We update the entity and relation embeddings using the update function over the smoothed enclosing subgraph. Subsequently, we readout the smoothed subgraph and get a global semantic representation $\mathbf{h}_{sem}$. In the process of structure refining, we emphasize the interaction structure of $g$ and refine it as $\tilde{g}$ using graph structure learning. Upon obtaining the refined $\tilde{g}$, the GNN is used to update the node embedding, and a max pooling operation is adopted to readout the global structure representation $\mathbf{h}_{str}$. We input the smooth semantic and structural embeddings of the enclosing subgraph into a classifier and output the interaction probability $p_{(u,r,v)}$. Finally, we calculate the loss of given batch samples and update the parameters $\Theta$ using gradient descent (i.e., the Adam~\citep{kingma2014adam} optimizer is used).

\begin{algorithm}[t]
\SetAlgoLined
\SetAlgoNoEnd
\SetKwInOut{Data}{Input}
\SetKwInOut{Result}{Output}
\Data{Enclosing subgraph size $k$;
Enclosing subgraph extracting function $F_{g}$;
Knowledge graph $\mathcal{G}$;
The number of iterations $epoch$;
}
\Result{A trained S$^2$DN model.}
Initialize entity embedding $\mathbf{X}$ by designed node features (Appendix B.1.3)\;
Initialize relation embedding $\mathbf{E}$ by Xavier initializer~\citep{glorot2010understanding}\;
\For{$i=1\rightarrow epoch$}
{
  \For{$((u,r,v), y_{(u,r,v)})\in \mathbf{U}$}{
  Extract enclosing subgraph $g$ using $F_{g}(\mathcal{G},(u,r,v))$\;
  Smooth relational semantic of relations within $g$ to get blurred relations $\tilde{R}$ (Eq. (2))\;
  Get smoothed relation embedding $\tilde{R}$ (Eq. (2))\;
  Refine the structure of $g$ as $\tilde{g}$ (Eqs. (5), (6), and (7))\;
  Readout the smoothed semantic graph $g$ (Eqs. (3) and (4)) and refined graph $\tilde{g}$ (Eq. (8)) to representation $\mathbf{h}_{sem}$ and $\mathbf{h}_{str}$\;
  $p_{(u,r,v)}$ $\gets$ $Classifier([\mathbf{h}_{sem};\mathbf{h}_{str}])$ (Eq. (11))\;
  $\ell(u,r,v)$ $\gets$ $Loss(p_{(u,r,v)}, y_{(u,r,v)})$ (Eq. (12))\;
  Update the trainable parameters $\Theta$ using gradient descent\;
  }
}
\textbf{return} $\mathcal{F}(\cdot|(\Theta, \mathcal{G}, g))$ \;
\caption{Semantic-aware Denoising Network}
\label{algo:full_process}
\end{algorithm}

\subsection{A.3 Computational Efficiency of S$^2$DN}
S$^2$DN consists of two main modules: Semantic Smoothing and Structure Refining. The semantic smoothing module contains an RGNN model over subgraphs whose computational complexity is $kD+\frac{kD^2}{2}+(kD)\cdot d\cdot L$, where $k$ is the subgraph size, $D$ denotes the average degree of KGs, $d$ is the embedding dimension, $kD$ represents the average number of nodes within the subgraph, and $L$ is the number of RGNN layers. The structure refining module includes a graph structure learning model and a GNN model to refine and represent the enclosing subgraphs, whose efficiency is $kD^2\cdot d\cdot L$. The overall computational complexity of S$^2$DN is $kD(((k+L)D+L+1)+\frac{D}{2}+1)$. This shows the complexity of the S$^2$DN approximation polynomial, and its efficiency depends mainly on the subgraph size, the average degree of KG, and the embedding dimension. Therefore, S$^2$DN can be scaled to large-scale KGs with small node degrees, and the training efficiency can be balanced for large-scale datasets by reducing the embedding dimension and decreasing the size of enclosing subgraphs.

\begin{figure}
\centering
    \includegraphics[width=0.85\columnwidth]{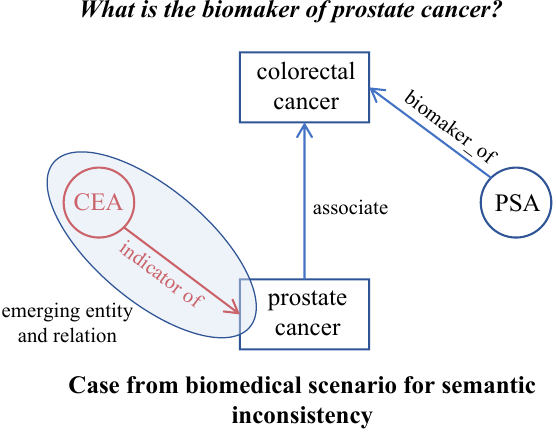} 
    \caption{An example to show the impact of semantic inconsistency.}
    \label{fig:cases_intro}
\end{figure}
\subsection{A.4 Cases of Semantic Inconsistency}
We show a case from the biomedical scenario to highlight the negative impact of semantic inconsistency in the reasoning process. As shown in Figure~\ref{fig:cases_intro}, for the query ``\textit{What is the biomarker of prostate cancer}?'', we observe that the answer will be misleading due to the semantic of emerging \textit{indicator\_of} is limited. This phenomenon suggests that semantic consistency plays an important role in reasoning for inductive scenarios.

\section{B. Experimental Settings}\label{app:imple_exp}
The experiments were carried out using both GeForce RTX 2080Ti and GeForce RTX 3090. S$^2$DN was implemented in PyTorch, and each experiment was repeated five times to ensure the reliability of the results.

\begin{table}
\centering
\caption{The hyperparameter details of S$^2$DN on all datasets.}
\label{tab:hyper_param}
\resizebox{1\columnwidth}{!}{%
\begin{tabular}{l|rrr}
\toprule
\textbf{Hyperparameters} & \multicolumn{1}{l}{\textbf{WN18RR}} & \multicolumn{1}{l}{\textbf{FB15k-237}} & \multicolumn{1}{l}{\textbf{NELL-995}} \\ \midrule
$dim$                      & 64                                  & 64                                     & 64                                    \\
$k$-hop subgraph           & 4                                   & 3                                      & 2                                     \\
learning rate            & 0.1                               & 0.0005                                   & 0.001                                 \\
batch size               & 8                                   & 32                                     & 8                                     \\
$\pi$                       & 0.5                                 & 0.5                                    & 0.5                                   \\
$\lambda$                   & 0.5                                 & 0.1                                    & 0.5      \\ 
$F(\cdot,\cdot)$                   & Attention                                 & Attention                                    & Attention \\ \bottomrule                            
\end{tabular}%
}
\caption{The hyperparameter details of S$^2$DN on all datasets.}
\label{tab:hyper_param}
\end{table}

\subsection{B.1 Implementation Details}
\subsubsection{B.1.1 Evaluation Protocol} \label{app:evaluate}
We adhere to the same inductive settings as GraIL~\citep{teru2020inductive}, where the training KG is denoted as $\mathcal{G}_{tra}=\{\mathcal{V}_{tra},\mathcal{R},\mathcal{F}_{tra}\}$, and the testing KG is denoted as $\mathcal{G}_{tst}=\{\mathcal{V}_{tst},\mathcal{R},\mathcal{F}_{tst}\}$. The relations remain consistent between training and testing, with disjoint sets of entities. We utilize three sets of triples represented as $\mathcal{T}_{tra}/\mathcal{T}_{val}/\mathcal{T}_{tst}$, which include reverse relations. $\mathcal{F}_{tra}$ is employed for training $\mathcal{T}_{tra}$ and validation $\mathcal{T}_{val}$, respectively. During the testing phase, $\mathcal{F}_{tra}$ is utilized to predict $\mathcal{T}_{tst}$ for evaluation. Similar to RED-GNN~\citep{zhang2022knowledge} and Adaprop~\citep{zhang2023adaprop}, our focus lies on the ranking performance of various methods. We adopt the filtered ranking metrics \textbf{Hits@1}, \textbf{Hits@10}, and mean reciprocal rank (\textbf{MRR}) as our evaluation metrics, where larger values indicate superior performance. We
follow GraIL and RMPI to rank each answer tail (or head) entity against 50
randomly sampled negative entities, rather than all negative entities in AdaProp.

\begin{table}
\centering

\resizebox{1\columnwidth}{!}{%
\begin{tabular}{cl|lll|lll|lll} \toprule
\multirow{2}{*}{\textbf{Version}} & \multicolumn{1}{c}{\multirow{2}{*}{\textbf{Type}}} & \multicolumn{3}{c}{\textbf{WN18RR}}                                                                    & \multicolumn{3}{c}{\textbf{FB15k-237}}                                                                 & \multicolumn{3}{c}{\textbf{NELL-995}}                                                                  \\
                                  & \multicolumn{1}{c}{}                               & \multicolumn{1}{c}{\textbf{\#R}} & \multicolumn{1}{c}{\textbf{\#N}} & \multicolumn{1}{c}{\textbf{\#E}} & \multicolumn{1}{c}{\textbf{\#R}} & \multicolumn{1}{c}{\textbf{\#N}} & \multicolumn{1}{c}{\textbf{\#E}} & \multicolumn{1}{c}{\textbf{\#R}} & \multicolumn{1}{c}{\textbf{\#N}} & \multicolumn{1}{c}{\textbf{\#E}} \\ \bottomrule
\multirow{2}{*}{\textbf{V1}}      & \textbf{TR}                                        & 9                                & 2,746                             & 6,678                             & 183                              & 2,000                             & 5,226                             & 14                               & 10,915                            & 5,540                             \\
                                  & \textbf{TE}                                        & 9                                & 922                              & 1,991                             & 146                              & 1,500                             & 2,404                             & 14                               & 225                              & 1,034                             \\ \midrule
\multirow{2}{*}{\textbf{V2}}      & \textbf{TR}                                        & 10                               & 6,954                             & 18,968                            & 203                              & 3,000                             & 12,085                            & 88                               & 2,564                             & 10,109                            \\
                                  & \textbf{TE}                                        & 10                               & 2,923                             & 4,863                             & 176                              & 2,000                             & 5,092                             & 79                               & 4,937                             & 5,521                             \\ \midrule
\multirow{2}{*}{\textbf{V3}}      & \textbf{TR}                                        & 11                               & 12,078                            & 32,150                            & 218                              & 4,000                             & 22,394                            & 142                              & 4,647                             & 20,117                            \\
                                  & \textbf{TE}                                        & 11                               & 5,084                             & 7,470                             & 187                              & 3,000                             & 9,137                             & 122                              & 4,921                             & 9,668                             \\ \midrule
\multirow{2}{*}{\textbf{V4}}      & \textbf{TR}                                        & 9                                & 3,861                             & 9,842                             & 222                              & 5,000                             & 33,916                            & 77                               & 2,092                             & 9,289                             \\
                                  & \textbf{TE}                                        & 9                                & 7,208                             & 15,157                            & 204                              & 3,500                             & 14,554                            & 61                               & 3,294                             & 8,520   \\ \bottomrule                         
\end{tabular}%
}%
\caption{Statistics of inductive benchmark datasets.}
\label{tab:data_stat}
\end{table}

\subsubsection{B.1.2 Hyper-parameter Settings} \label{appdix:hyper_params}
We adopt the Xavier initializer~\citep{glorot2010understanding} to initialize the model parameters and optimize S$^2$DN with Adam~\citep{kingma2014adam}. The grid search is applied to retrieve the best hyperparameters. For enclosing subgraph extraction, we set the size of subgraphs as $k=4$ (i.e., 4-hop subgraphs) for WN18RR, $k=3$ for FB15k-237, and $k=2$ for NELL-995. We set the embedding size $dim=64$ for all datasets, learning rate $lr=0.1$ for WN18RR, $lr=0.0005$ for FB15K-237, and $lr=0.001$ for NELL-995. We tune the pruning threshold $\pi$ in $\{0.1, 0.3, 0.5, 0.7, 0.9\}$ and select $\pi=0.5$ for all datasets.
We show the details of hyperparameters in Table~\ref{tab:hyper_param}.

\subsubsection{B.1.3 Details of Node Feature} \label{appdix:node_feat}
Following GraIL,
to capture the global semantic representations of enclosing subgraphs surrounding the target link, S$^2$DN employs RGNN to represent it. RGCN necessitates node features for initializing the message passing algorithm. 
We extend their double radius node labeling scheme proposed by~\citep{zhang2018link,teru2020inductive} to our framework. 
For a given link $(u,r,v)$, each node in the subgraph surrounding the target link is featured by the pair $(d(i,u),d(i,v))$, where $d(\cdot,\cdot)$ indicates the shortest distance between the input nodes. This procedure extracts the positional information for each node, reflecting its structural role within the target subgraph. The node $u$ and $v$ of the link are uniquely represented as $(0,1)$ and $(1,0)$, respectively, which can be identified by S$^2$DN. Subsequently, the node features are integrated as [one-hot($d(i,u)$)$\oplus$one-hot($d(i,v)$)], where $\oplus$ denotes concatenation operation.
% \vspace{-0.5em}

\begin{table}[t]
% \begin{minipage}{0.5\textwidth}

\centering
\resizebox{1\columnwidth}{!}{%
\begin{tabular}{l|l|cccc} \toprule
\textbf{Noise Type}                 & \textbf{Model} & \multicolumn{1}{l}{\textbf{0\%}} & \textbf{15\%} & \textbf{35\%} & \textbf{50\%} \\ \midrule
\multirow{4}{*}{\textbf{Semantic}}  & RMPI                & $61.50$                          & $59.88_{{ 2.6\%}}$           & $57.56_{{ 6.4\%}}$         & $55.39_{{ 9.9\%}}$         \\
                                    & S$^2$DN w/o SS          & $57.00$                          & $55.98_{{ 1.7\%}}$         & $54.21_{{ 4.8\%}}$         & $52.89_{{ 7.2\%}}$         \\
                                    & S$^2$DN w/o SR          & $58.50$                          & $57.45_{{ 1.7\%}}$         & $56.09_{{ 4.1\%}}$         & $54.79_{{ 6.3\%}}$         \\
                                    & S$^2$DN                 & $63.50$                          & $62.58_{{ 1.4\%}}$         & $61.87_{{ 2.5\%}}$         & $60.79_{{ 4.2\%}}$         \\ \midrule
\multirow{4}{*}{\textbf{Structure}} & RMPI                & $61.50$                          & $60.46_{{ 1.7\%}}$         & $58.33_{{ 5.2\%}}$         & $56.24_{{ 8.6\%}}$         \\
                                    & S$^2$DN w/o SS          & $57.00$                          & $55.67_{{ 2.3\%}}$         & $54.87_{{ 3.7\%}}$         & $53.15_{{ 6.7\%}}$         \\
                                    & S$^2$DN w/o SR          & $58.50$                          & $56.47_{{ 3.5\%}}$         & $54.49_{{ 6.8\%}}$         & $53.01_{{ 9.3\%}}$         \\
                                    & S$^2$DN                 & $63.50$                          & $62.55_{{ 1.4\%}}$         & $61.86_{{ 2.5\%}}$         & $60.99_{{ 3.9\%}}$   \\ \bottomrule     
\end{tabular}%
}
\caption{The results (\textbf{Hits@10}) of S$^2$DN on \textbf{NELL-995\_V1} under different noise ratios. The blue subscripts represent the rate of performance decline over contaminated KGs.}
\label{tab:robust_nell}
% \end{minipage}
\end{table}

\subsubsection{B.1.4 Datasets}\label{appdx:dataset}
In this study, we utilize three widely-used datasets: WN18RR~\citep{dettmers2018convolutional}, FB15k-237~\citep{toutanova2015representing}, and NELL-995~\citep{xiong2017deeppath}, to evaluate the performance of S$^2$DN and baseline models. Following~\citep{teru2020inductive,zhang2023adaprop}, we use the same four subsets with increasing size of the three datasets, resulting in a total of 12 subsets. Each subset comprises distinct training and test sets. Table~\ref{tab:data_stat} presents the detailed statistics of the datasets.

\subsection{B.2 Baselines}\label{app:baseline}
\subsubsection{B.2.1 Details of Baseline Methods} 
% \subsubsection{Baseline Methods}
To verify the performance of S$^2$DN, we compare it against various state-of-the-art baselines from Rule- and GNN-based perspectives as follows:
\begin{itemize}
    \item \textbf{Rule-based Methods}: \textbf{NeuralLP}~\citep{yang2017differentiable} and \textbf{DRUM}~\citep{sadeghian2019drum} are embedding-free models that learn logical rules from knowledge graphs for inductive link prediction with unseen entities. On the other hand, \textbf{A$^*$Net}~\cite{zhu2024net} is a scalable path-based method for knowledge graph reasoning.
    % \textbf{RuleN}~\citep{meilicke2018fine} combines the explanatory qualities of rule-based systems with the precision of embedding-based models for high-quality knowledge graph completion.
    \item \textbf{GNN-based Methods}: \textbf{GraIL}~\citep{teru2020inductive} employs reasoning over enclosing subgraph structures to capture a robust inductive bias for learning entity-independent semantics. Meanwhile,
    \textbf{CoMPILE}~\citep{mai2021communicative} designs a communicative message passing network to effectively handle complex relations. Additionally, \textbf{TACT}~\citep{chen2021topology} incorporates entity-based message passing and relational correlation modules to model topological relations. \textbf{SNRI}~\citep{xu2022subgraph} utilizes subgraph neighboring relations infomax to exploit relational paths, while \textbf{RMPI}~\citep{geng2023relational} proposes a novel relational message passing network to make fully leverage relation patterns for subgraph reasoning. 
\end{itemize}

\begin{table*}[t]
\centering
\resizebox{0.9\textwidth}{!}{%
\begin{tabular}{l|lll|lll|lll|lll} \toprule
\multicolumn{1}{c|}{\multirow{2}{*}{\textbf{Methods}}} & \multicolumn{3}{c}{\textbf{V1}}                  & \multicolumn{3}{c}{\textbf{V2}}                  & \multicolumn{3}{c}{\textbf{V3}}                  & \multicolumn{3}{c}{\textbf{V4}}                  \\ \cline{2-13} 
\multicolumn{1}{c|}{}                                  & Hits@1         & Hits@10        & MRR            & Hits@1         & Hits@10        & MRR            & Hits@1         & Hits@10        & MRR            & Hits@1         & Hits@10        & MRR            \\ \midrule
DRUM                                                   & 10.50          & 19.54          & 12.28          & 51.71          & 78.47          & 60.65          & 51.62          & 82.71          & 62.75          & 43.63          & 80.85          & 54.73          \\
NeuralLP                                               & 19.05          & 40.78          & 26.16          & 46.49          & 78.73          & 58.61          & 52.67          & 82.18          & 61.67          & 40.37          & 80.58          & 53.61          \\
A$^*$Net                                                  & 38.45          & 56.50          & 43.15          & 60.04          & 81.15          & 68.35          & 54.03          & 83.57          & 61.81          & 37.48          & 75.33          & 46.33          \\ \midrule
CoMPILE                                                & \underline{46.50}          & 61.50          & 51.18          & 63.13          & 89.49          & 70.42          & 71.19          & 90.91          & \underline{79.08}          & 15.94          & 19.97          & 20.08          \\
TAGT                                                   & 45.00          & 58.00          & 50.13          & 55.57          & \underline{92.01}          & 68.18          & 59.39          & 93.88          & 71.44          & 43.64          & 81.33          & 57.47          \\
SNRI                                                   & 44.00          & 53.00          & 48.60          & 41.28          & 75.63          & 53.7           & 58.89          & \underline{94.07}          & 71.07          & 19.29          & 42.20          & 28.34          \\
RMPI                                                   & \textbf{47.50} & \underline{61.50}          & \underline{51.68}          & \underline{64.13}          & 91.81          & \underline{71.87}          & 71.69          & 92.89          & \textbf{80.07}          & \textbf{64.50}          & \textbf{84.20}          & \textbf{72.63}          \\ \midrule
S$^2$DN                                           & 45.00          & \textbf{63.50} & \textbf{52.05} & \textbf{64.35} & \textbf{93.27} & \textbf{72.89} & \textbf{76.87} & \textbf{94.34} & 76.22 & \underline{56.98} & \underline{74.89} & \underline{64.12} \\
S$^2$DN w/o SS                                    & 40.50                                                             & 57.00                                                                & 48.06                            & 61.45                                                            & 91.18                                                             & 72.75                            & \underline{73.08}                                                            & 90.21                                                             & 75.12                            & 54.24                                                            & 72.01                                                    & 63.11                   \\
S$^2$DN w/o SR                                    & 45.00                                                               & 58.50                                                              & 50.95                            & 65.55                                                            & 92.02                                                             & 75.46                            & 60.63                                                            & 91.34                                                             & 71.45                            & 20.66                                                            & 54.04                                                    & 30.74  
         \\ \bottomrule
\end{tabular}
}%
\caption{The performance (i.e., \textbf{Hits@1}, \textbf{Hits@10}, \textbf{MRR}, in percentage) of S$^2$DN on the NELL-995 dataset. The boldface denotes the highest score and the underline indicates the best baseline.}
\label{tab:nell_995}
\end{table*}

\subsubsection{B.2.2 Implementation of Baselines}
For rule-based methods, we follow GraIL\footnote{https://github.com/kkteru/grail} and implement them using their public code. A$^*$Net is implemented by using their official code\footnote{https://github.com/DeepGraphLearning/AStarNet}.
We reproduced the results of GraIL, CoMPILE\footnote{https://github.com/TmacMai/CoMPILE\_Inductive\_Knowledge\_Graph}, RMPI\footnote{https://github.com/zjukg/RMPI}, SNRI\footnote{https://github.com/Tebmer/SNRI}, and TACT\footnote{https://github.com/zjukg/RMPI/tree/main/TACT} based on their source code and public optimal hyper-parameters. Due to no NELL-995 data in the source of SNRI, we use the data from GraIL to evaluate SNRI.
\subsubsection{B.2.3 Missing Baseline Methods}
% We exclude ConGLR~\citep{lin2022incorporating} and CBGNN~\citep{yan2022cycle} from our baseline methods as they have the same experimental setting as GraIL by sampling 50 negative candidates for each query. 
ConGLR~\citep{lin2022incorporating} and CBGNN~\citep{yan2022cycle} lack the implementation for expanding the candidate set in their source code. Additionally, RED-GNN~\citep{zhang2022knowledge} and AdaProp~\citep{zhang2023adaprop} differ in evaluation settings from GraIL, as they sample all entities for each query, making direct adoption challenging. Consequently, we did not consider them as baseline methods.

\section{C. Additional Experiments}
\subsection{C.1 Inductive KGC performance of S$^2$DN on NELL-995}\label{ind_kgc_nell}

The inductive KGC performance of S$^2$DN on the NELL-995 dataset is shown in Table~\ref{tab:nell_995}. We can observe that S$^2$DN achieves comparable performance with previous rule- and GNN-based models on the NELL-995 dataset.

% \begin{figure*}[t]
% % \begin{wrapfigure}{r}{0.4\textwidth}
% \centering
% \subfigure[WN18RR]{
% \centering
%     \includegraphics[width=0.28\textwidth]{figs/wn18.pdf} 
% }
% \subfigure[FB15k-237]{
%     \centering
%     \includegraphics[width=0.28\textwidth]{figs/fb237.pdf} 
% }
% \subfigure[NELL-995]{
%     \centering
%     \includegraphics[width=0.28\textwidth]{figs/nell995.pdf} 
% }
% \caption{The transition ratio between the original and blurred relations on three datasets (V1 version). The element $m_{ij}$ in the matrix represents the proportion of the relation $i$ is smoothed to relation $j$.}
% \label{fig:heatmap}
% % \end{wrapfigure}
% \end{figure*}

\begin{figure*}

\centering
\subfigure[GraIL, WN18RR (q1)]{
\centering
    \includegraphics[width=0.22\textwidth]{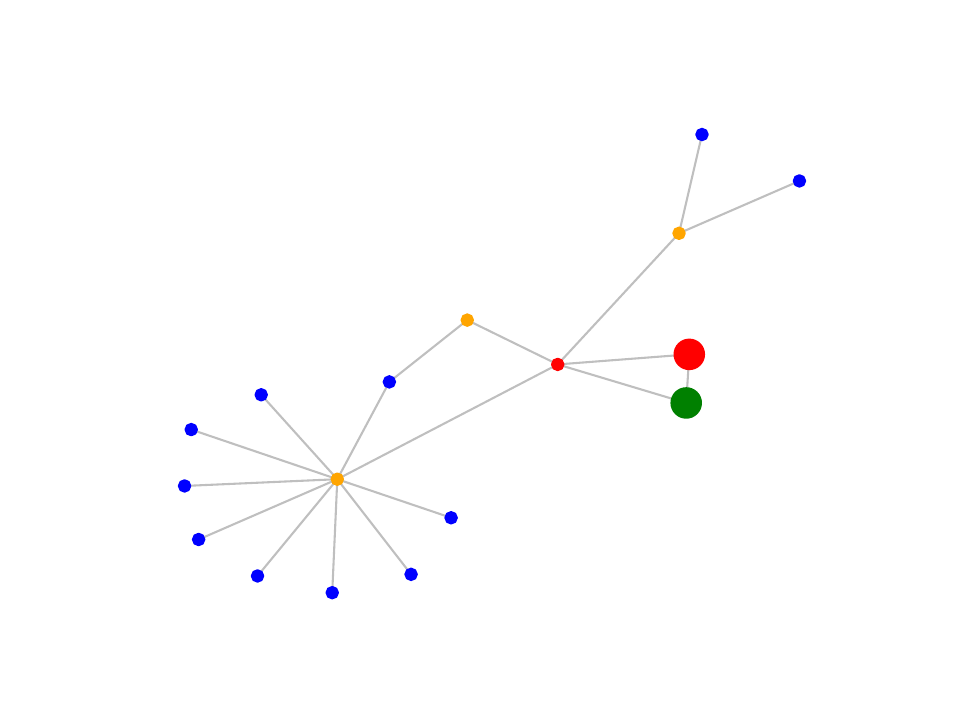} 
}
\subfigure[S$^2$DN, WN18RR (q1)]{
    \centering
    \includegraphics[width=0.22\textwidth]{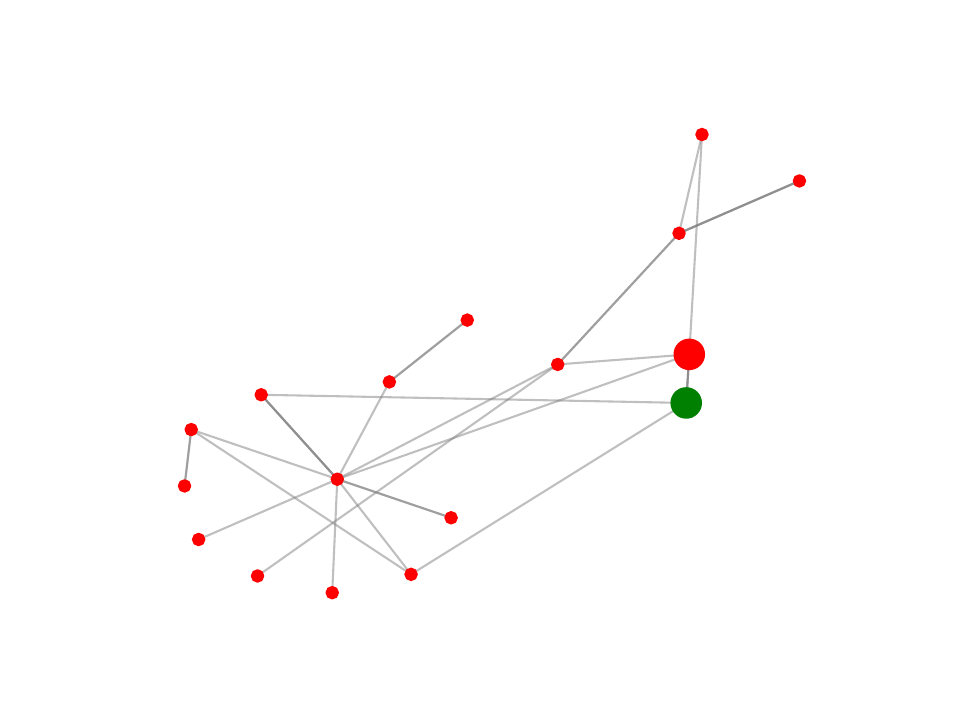} 
}
\subfigure[GraIL, WN18RR (q2)]{
\centering
    \includegraphics[width=0.22\textwidth]{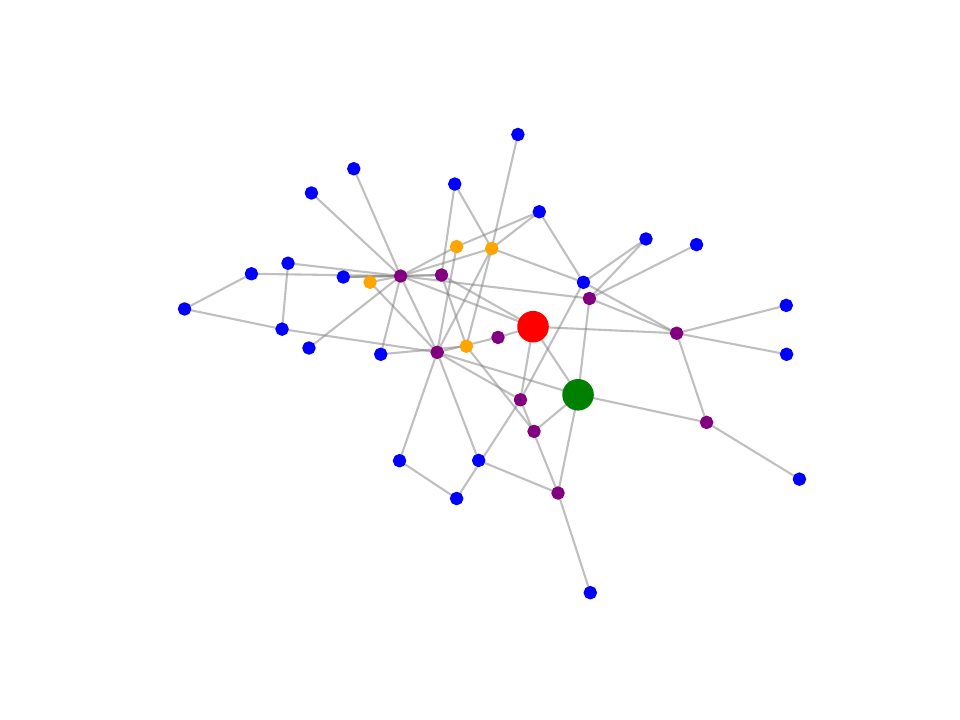} 
}
\subfigure[S$^2$DN, WN18RR (q2)]{
    \centering
    \includegraphics[width=0.22\textwidth]{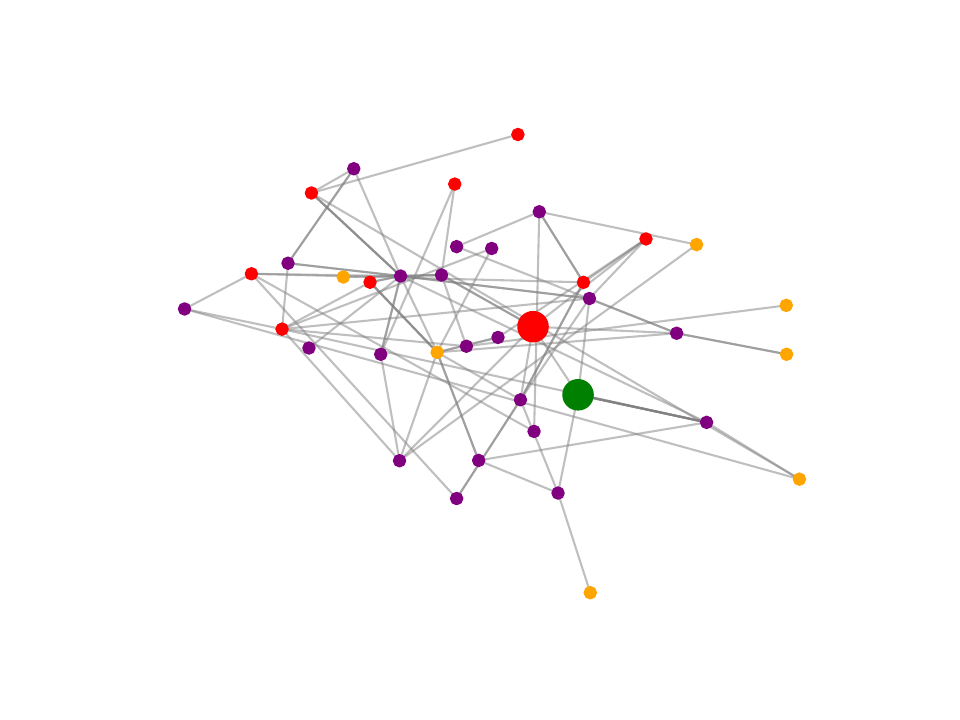} 
}

% \subfigure[GraIL, FB15k (q1)]{
% \centering
%     \includegraphics[width=0.22\textwidth]{figs/fb_q1_original.pdf} 
% }
% \subfigure[S$^2$DN, FB15k (q1)]{
%     \centering
%     \includegraphics[width=0.22\textwidth]{figs/fb_q1_learned.pdf} 
% }
% \subfigure[GraIL, FB15k (q2)]{
% \centering
%     \includegraphics[width=0.22\textwidth]{figs/fb_q2_original.pdf} 
% }
% \subfigure[S$^2$DN, FB15k (q2)]{
%     \centering
%     \includegraphics[width=0.22\textwidth]{figs/fb_q2_learned.pdf} 
% }

\subfigure[GraIL, NELL (q1)]{
\centering
    \includegraphics[width=0.22\textwidth]{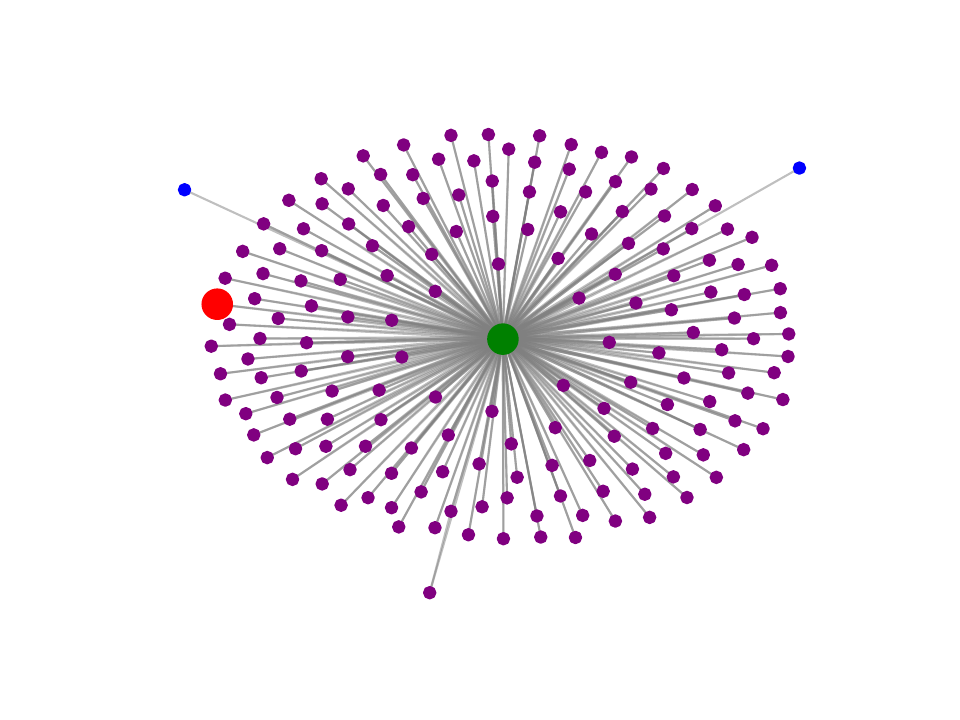} 
}
\subfigure[S$^2$DN, NELL (q1)]{
    \centering
    \includegraphics[width=0.22\textwidth]{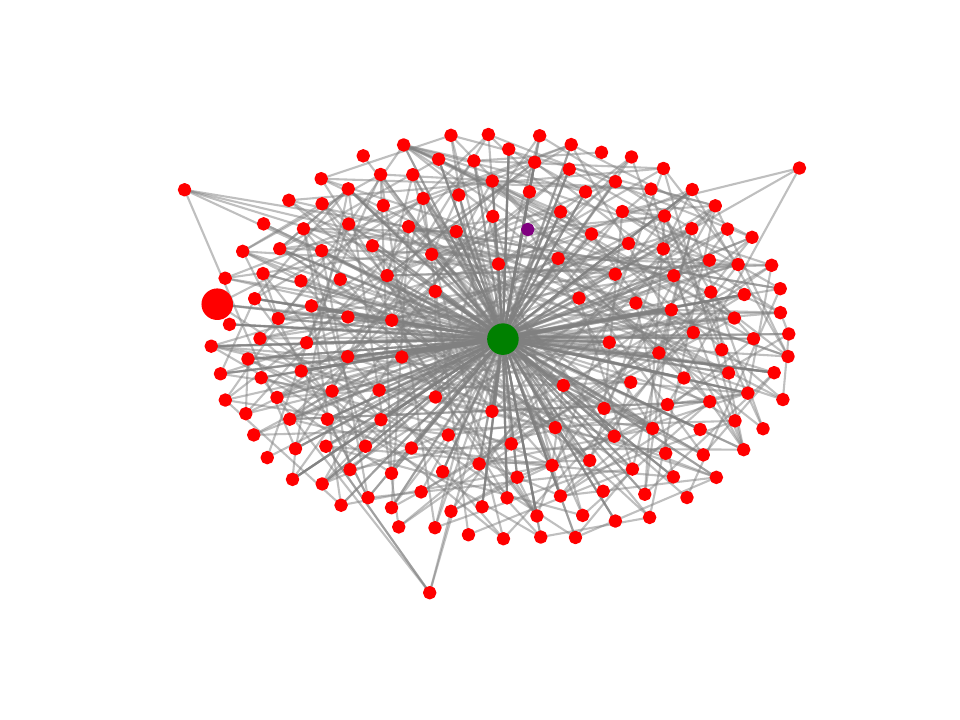} 
}
\subfigure[GraIL, NELL (q2)]{
\centering
    \includegraphics[width=0.22\textwidth]{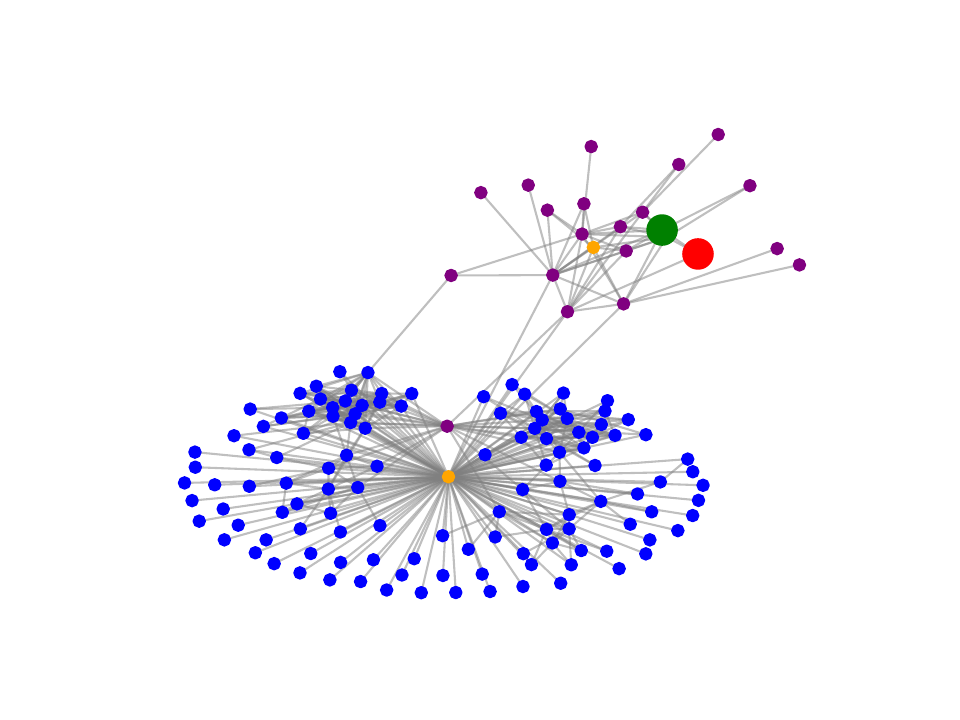} 
}
\subfigure[S$^2$DN, NELL (q2)]{
    \centering
    \includegraphics[width=0.22\textwidth]{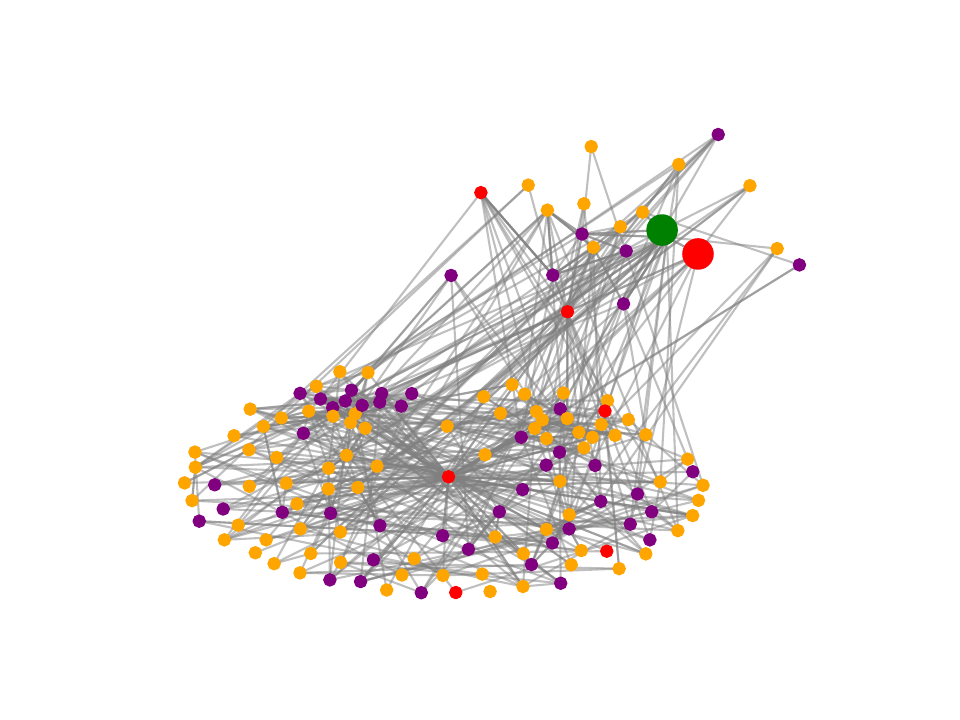} 
}
\caption{The big red and green nodes represent the source and target entities. The small nodes in red, orange, and blue are shared entities involved in the $1\sim 3$-hop between source and target nodes. The nodes in purple indicate unshared entities.}
\label{fig:hetmap}
% % \end{figure}
% \end{wrapfigure}
\end{figure*}

\subsection{C.2 Reliability of S$^2$DN on NELL-995}\label{app:robust_nell}
We generate different proportions of \textit{semantic} and \textit{structural} negative interactions (i.e., 5\%, 15\%, 35\%, and 50\%) to contaminate the training KGs.
We show the performance of RMPI, S$^2$DN, and its variants on noisy KGs of the NELL-995 dataset in Table~\ref{tab:robust_nell}. As the conclusion of Section~\ref{sec:robust_exp}, we can observe the same phenomenon in NELL-995. This observation shows S$^2$DN can effectively mitigate unconvincing knowledge while providing reliable local structure.

% \subsection{C.3 Relation Summary}

\subsection{C.3 Visualization}
To explicitly demonstrate the ability of S$^2$DN to provide reliable links to downstream tasks, we designed case studies on three datasets. We visualize the exemplar reasoning subgraph of S$^2$DN (i.e., the refined subgraph) and GraIL (i.e., the original subgraph) models for different queries (selected from the V1 version of all three datasets, and ) in Figure~\ref{fig:hetmap}. As illustrated in Figure~\ref{fig:hetmap}, we observe that compared to GraIL, S$^2$DN can provide more knowledge for enhanced subgraph reasoning while retaining the original reliable information. For example, Figure~\ref{fig:hetmap}(a) and Figure~\ref{fig:hetmap}(b) show the subgraphs from GraIL and S$^2$DN have a similar layout, while S$^2$DN offers more links and filter out irrelevant interaction between source and target entities. This indicates that S$^2$DN is effective in subgraph reasoning inductively by a structure-refined mechanism.

% \begin{wrapfigure}{r}{0.5\textwidth}

\begin{figure*}[t]
\centering
\includegraphics[width=1\textwidth]{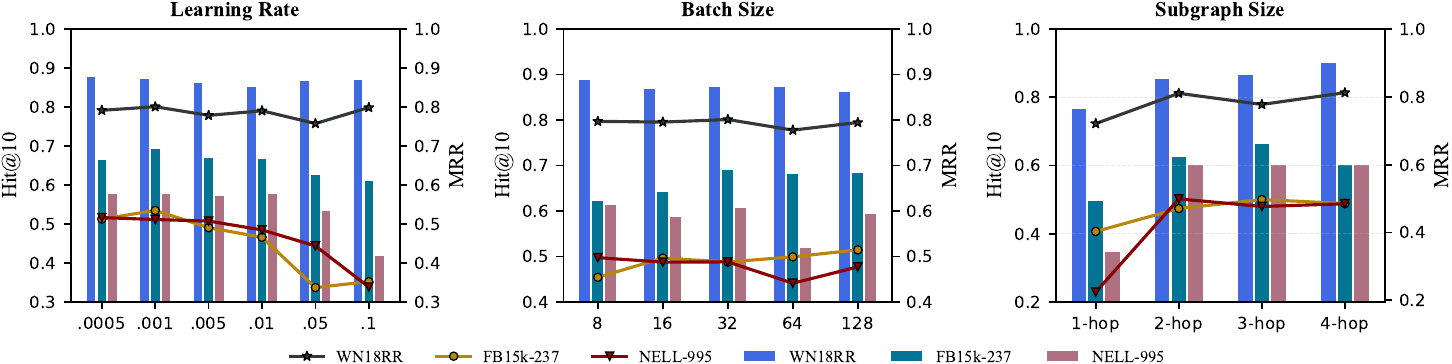} % Reduce the figure size so that it is slightly narrower than the column. Don't use precise values for figure width. This setup will avoid overfull boxes.
\caption{The sensitivity of hyperparameters across all datasets (V1 version). The bar indicates the \textbf{Hits@10} performance and the line denotes the \textbf{MRR} results.}
\label{fig:param_anay}
\end{figure*}

% \subsection{Robustness of S$^2$DN on NELL-995}

\subsection{C.4 Hyperparameter Sensitivity}
This section focuses on assessing the influence of different hyperparameter configurations on the inductive KGC task. To accomplish this, we conduct a thorough hyperparameter analysis using the V1 versions of three datasets: WN18RR\_V1, FB15k-237\_V1, and NELL-995\_V1.

\subsubsection{C.4.1 Impact of learning rate}
We investigate the effect of the learning rate of S$^2$DN by varying it from 0.0005 to 0.1 over three datasets. As illustrated in Figure~\ref{fig:param_anay}, we can find that using S$^2$DN with a lower learning rate (e.g., 0.0005 for WN18RR and NELL-995, and 0.001 for FB15k-237) performs better on three datasets than a higher one. Specifically, as the learning rate increases, the performance on the FB15k-237 and NELL-995 decreases, indicating that S$^2$DN is sensitive to the setting of the learning rate over these two datasets. In contrast, the performance of S$^2$DN on WN18RR is stable on the WN18RR dataset, so we can set a higher learning rate to accelerate the convergence of the model. As a result, we set the learning rate of S$^2$DN on WN18RR, FB15k-237, and NELL-995 as 0.1, 0.0005, and 0.001 respectively.

\subsubsection{C.4.2 Impact of batch size}
We experiment with different batch sizes ranging from 8 to 128. As depicted in the central portion of Figure~\ref{fig:param_anay}, the optimal performance of S$^2$DN is observed with a batch size of 8 for WN18RR \& NELL-995 and 32 for FB15k-237. We set the default batch sizes as 8 for WN18RR \& NELL-995 and 32 for FB15k-237.

\subsubsection{C.4.3 Impact of subgraph size}
To assess the efficacy of S$^2$DN across various subgraph sizes, we conduct experiments to explore the impact of $k$-hop subgraphs on predictive performance. We investigate subgraph sizes ranging from $1$-hop to $4$-hop. Results indicate that S$^2$DN achieves optimal performance with $k=4$ for the WN18RR\_V1 and NELL-995\_V1, and $k=3$ for FB15k-237\_V1. Interestingly, the Hits@10 metric for S$^2$DN exhibits a hump-shaped trend across different subgraph sizes in the FB15k-237\_V1 dataset. This phenomenon suggests that while an adequate subgraph size can provide valuable information, excessively large sizes may introduce noise. In contrast, S$^2$DN performs better on WN18RR\_V1 and NELL-995\_V1 with larger subgraph sizes, implying that such sizes can enhance performance by incorporating more positive information, particularly on sparse graphs (e.g., WN18RR and NELL-995 with lower average degrees compared to FB15k). Consequently, we designate the subgraph sizes for WN18RR, FB15k-237, and NELL-995 as $4$-hop, $3$-hop, and $4$-hop.

\begin{table}
\caption{The performance (Hits@10) of S$^2$DN with different reliability estimators. The results are tested on the V1 version of all datasets.}
\label{tab:estimator}
\resizebox{1\columnwidth}{!}{%
\begin{tabular}{lrrr} \toprule
\textbf{Reliability Estimator} & \multicolumn{1}{l}{\textbf{WN18RR}} & \multicolumn{1}{l}{\textbf{FB15k-237}} & \multicolumn{1}{l}{\textbf{NELL-995}} \\ \midrule
Attention                      & \textbf{87.64}                               & \textbf{67.34}                                  & \textbf{64.11}                                 \\
MLP                            & 86.70                               & 65.61                                  & 63.21                                 \\
Weighted Cosine                & 87.50                               & 66.09                                  & 62.00     \\
Cosine                         & 87.20                               & 63.17                                  & 62.00                                 \\
 \bottomrule                           
\end{tabular}
}%
\caption{The performance (Hits@10) of S$^2$DN with different reliability estimators. The results are tested on the V1 version of all datasets.}
\label{tab:estimator}
\end{table}

\subsubsection{C.4.4 Impact of Reliability Estimation}\label{appdx:reliability}
We conduct experiments to investigate the impact of various reliability estimation functions $F(\cdot,\cdot)$ (defined in Section~\ref{sec:struc_refine}) by varying the estimation types to \textit{Attention}, \textit{MLP}, \textit{Weighted Cosine}, and \textit{Cosine}. As shown in Table~\ref{tab:estimator}, the \textit{Attention} with a linear attention mechanism modeling the reliability among the set of nodes in subgraphs achieves the best performance across all datasets. The \textit{MLP} adopts a 2-layer preceptor, which yields a secondary best result by learning reliable probability in the local subgraphs. This demonstrates the attention operation can better model the importance of node pairs, improving the effectiveness of estimating the reliable edges. In addition, the parametric \textit{Weighted\_Cosine} outperforms the non-parametric \textit{Cosine} showing the learned weight guided by downstream tasks is more efficient. Based on the above observations, we choose the \textit{Attention} as the reliability estimator of S$^2$DN.

\subsection{C.5 Comparison with LLaMA2-7B}
We use the LLM-based KGC model KAPING~\cite{baek2023knowledge} as our baseline. In KAPING, a query, its context (i.e., relevant facts surrounding the query), and its answer (e.g., the tail entity) are structured as fine-tuning instructions. These constructed instructions are then used to fine-tune an LLM, such as LLaMA2, enabling it to predict possible tail entities for given queries. The performance results, shown in Table~\ref{llama}, indicate that KAPING does not perform optimally on the WN18RR\_V1 and FB15k-237\_V1 datasets. This limitation is likely due to the lack of text information for entities (e.g., only identifier /m/0gq9h in FB15k-237 is given), which constrains the reasoning ability of LLMs. Additionally, we frame KAPING as a Yes/No question, like 'Is /m/01t\_vv the /film/film/genre of /m/0qf2t?'. KAPING can perform better with an accuracy of 69.81\%.

\begin{table}[t]
    \centering
    \begin{tabular}{lll}
    \hline
        ~ & WN18RR\_V1 & FB15k-237\_V1 \\ \hline
        KAPING (LLaMA2-7B) & 36.27 & 19.84 \\ 
        S2DN & 74.73 & 43.68 \\ \hline
    \end{tabular}
    \caption{The performance (Hits@1) of S$^2$DN and KAPING.}
    \label{llama}
\end{table}